\documentclass[acmsmall]{acmart}

\usepackage{amsmath}

\usepackage{subcaption}
\usepackage{multirow} 
\usepackage{algorithm}
\usepackage{algpseudocode}
\usepackage{microtype} 
\usepackage{array} 
\usepackage{booktabs} 
\usepackage{tcolorbox}
\usepackage{svg}
\usepackage{siunitx}

\usepackage{colortbl}

\newcommand{\RightComment}[1]{\hfill$\triangleright$\ \textit{#1}}
\algnewcommand{\Inputs}{\item[\bfseries Input:]}%
\algnewcommand{\Outputs}{\item[\bfseries Output:]}%
\newcolumntype{M}[1]{>{\centering\arraybackslash}m{#1}}
\AtBeginDocument{%
  }

\setcopyright{cc}
\setcctype{by-nc-nd}
\acmDOI{10.1145/3832183}
\acmYear{2026}
\acmJournal{PACMSE}
\acmVolume{3}
\acmNumber{ISSTA}
\acmArticle{ISSTA092}
\acmMonth{10}
\acmSubmissionID{issta26main-p847-p}
\received{2026-01-30}
\received[accepted]{2026-06-25}

\begin{document}

\title{Test Case Prioritization for DNNs via Neural Collapse Instability}


\author{Chunyu Liu}
\orcid{0009-0002-2230-3363}
\affiliation{%
  \institution{State Key Laboratory of Networking and Switching Technology, Beijing University of Posts and Telecommunications}
  \city{Beijing}
  \country{China}
}
\affiliation{%
  \institution{National Engineering Research Center of Disaster Backup and Recovery, Beijing University of Posts and Telecommunications}
  \city{Beijing}
  \country{China}
}
\email{chunyuliu@bupt.edu.cn}

\author{Mingyuan Li}
\orcid{0000-0003-2522-8426}
\affiliation{%
  \institution{Information Security Center, Beijing University of Posts and Telecommunications}
  \city{Beijing}
  \country{China}
}
\affiliation{%
  \institution{State Key Laboratory of Networking and Switching Technology, Beijing University of Posts and Telecommunications}
  \city{Beijing}
  \country{China}
}
\email{henryli\_i@bupt.edu.cn}

\author{Yang Li}
\orcid{0009-0006-3812-0448}
\affiliation{%
  \institution{State Key Laboratory of Networking and Switching Technology, Beijing University of Posts and Telecommunications}
  \city{Beijing}
  \country{China}
}
\email{liyang02@bupt.edu.cn}

\author{Wenmin Li}
\authornote{Corresponding author.}
\orcid{0000-0002-4665-8508}
\affiliation{%
  \institution{State Key Laboratory of Networking and Switching Technology, Beijing University of Posts and Telecommunications}
  \city{Beijing}
  \country{China}
}
\email{liwenmin@bupt.edu.cn}

\author{Fei Gao}
\orcid{0000-0002-1546-4364}
\affiliation{%
  \institution{State Key Laboratory of Networking and Switching Technology, Beijing University of Posts and Telecommunications}
  \city{Beijing}
  \country{China}
}
\affiliation{%
  \institution{National Engineering Research Center of Disaster Backup and Recovery, Beijing University of Posts and Telecommunications}
  \city{Beijing}
  \country{China}
}
\email{gaof@bupt.edu.cn}

\author{Tengfei Tu}
\orcid{0000-0001-5683-5347}
\affiliation{%
  \institution{State Key Laboratory of Networking and Switching Technology, Beijing University of Posts and Telecommunications}
  \city{Beijing}
  \country{China}
}
\email{tutengfei.kevin@bupt.edu.cn}

\author{Su-Juan Qin}
\orcid{0000-0002-6405-6711}
\affiliation{%
  \institution{State Key Laboratory of Networking and Switching Technology, Beijing University of Posts and Telecommunications}
  \city{Beijing}
  \country{China}
}
\email{qsujuan@bupt.edu.cn}









\renewcommand{\shortauthors}{
  Chunyu Liu, Mingyuan Li, Yang Li, Wenmin Li, Fei Gao, Tengfei Tu, and Su-Juan Qin
}


\begin{abstract}
With the widespread deployment of deep neural networks (DNNs) in safety-critical domains, reducing the cost of model validation under limited testing budgets has become increasingly important. 
Existing test case prioritization techniques often rely on single-checkpoint confidence signals derived from output probabilities.
However, DNNs can be confidently wrong, and the confidence margin between the predicted and competing classes is frequently small, which weakens early fault discovery. 
To address this limitation, we propose a \textbf{N}eural-\textbf{C}ollapse-\textbf{I}nspired \textbf{P}rioritization (\textbf{NCIP}) framework that replaces absolute confidence with cross-checkpoint prediction variability in the terminal training regime, where model geometry becomes highly structured.
NCIP introduces two key components. First, it selects an NC-guided representative subset of training checkpoints using an equiangularity score of classifier weights, quantified as the standard deviation of pairwise cosine similarities among class weight vectors. Second, it prioritizes test inputs by their prediction variability across the selected checkpoints, surfacing boundary-adjacent and failure-prone samples that are unstable under checkpoint-induced decision boundary shifts. 
Extensive experiments across multiple datasets and architectures show that NCIP achieves strong performance in early fault discovery compared with competitive baselines, with \textbf{\SIrange[range-phrase=--]{1.5}{16.6}{\percent}} RAUC-ALL gains and \textbf{\SIrange[range-phrase=--]{4.9}{20.6}{\percent}} RAUC-$500$ gains under the same testing budget. NCIP further attains the best average performance across all dataset-model pairs.
\end{abstract}

\ccsdesc[500]{Software and its engineering~Software testing and debugging}

\keywords{Test Case Prioritization, Deep Learning, Neural Collapse}


\maketitle

\section{Introduction}
Deep neural networks (DNNs) have revolutionized various industries by excelling at complex pattern recognition and decision-making tasks. However, as these technologies are increasingly adopted in safety-critical domains such as autonomous driving and medical diagnostics, their inherent vulnerabilities have become more apparent. The severe consequences of potential failures in these applications, ranging from fatal traffic accidents \cite{stewart2018tesla} to life-threatening medical misdiagnoses, highlight the urgent need for robust test frameworks.


As testing budgets are typically limited, an effective strategy is test case prioritization, which ranks a candidate test set so that faulty inputs are revealed as early as possible. A large body of work has studied prioritization criteria for DNN testing~\cite{li2024distance, shen2024prioritizing, hwang2023improved}, including coverage-driven heuristics~\cite{ma2018deepgauge, yin2025lightweight}, uncertainty-based scores~\cite{feng2020deepgini, bao2023defense, wang2025sets}, and learning-to-rank measures~\cite{wang2021prioritizing, dang2024test, shen2024prioritizing}. 


However, most test prioritization methods assume that confidence-based metrics can reliably separate correct from incorrect predictions, with correct predictions typically assigned higher confidence, whereas mispredictions exhibit low confidence. Recent studies~\cite{bao2023defense, chen2024fast} have examined this assumption, showing that DNNs can be confidently wrong and that the confidence gap between the predicted class and competing classes is often marginal even for correct cases. This confidence and accuracy mismatch affects prioritization: high-confidence errors may be ranked late, while low-confidence yet correct inputs may be prioritized excessively, consuming limited testing budgets.


\begin{figure}[tb]
  \centering
  \begin{tabular}{cc}
    \begin{subfigure}[b]{0.47\textwidth}
      \centering
      \includegraphics[width=\linewidth]{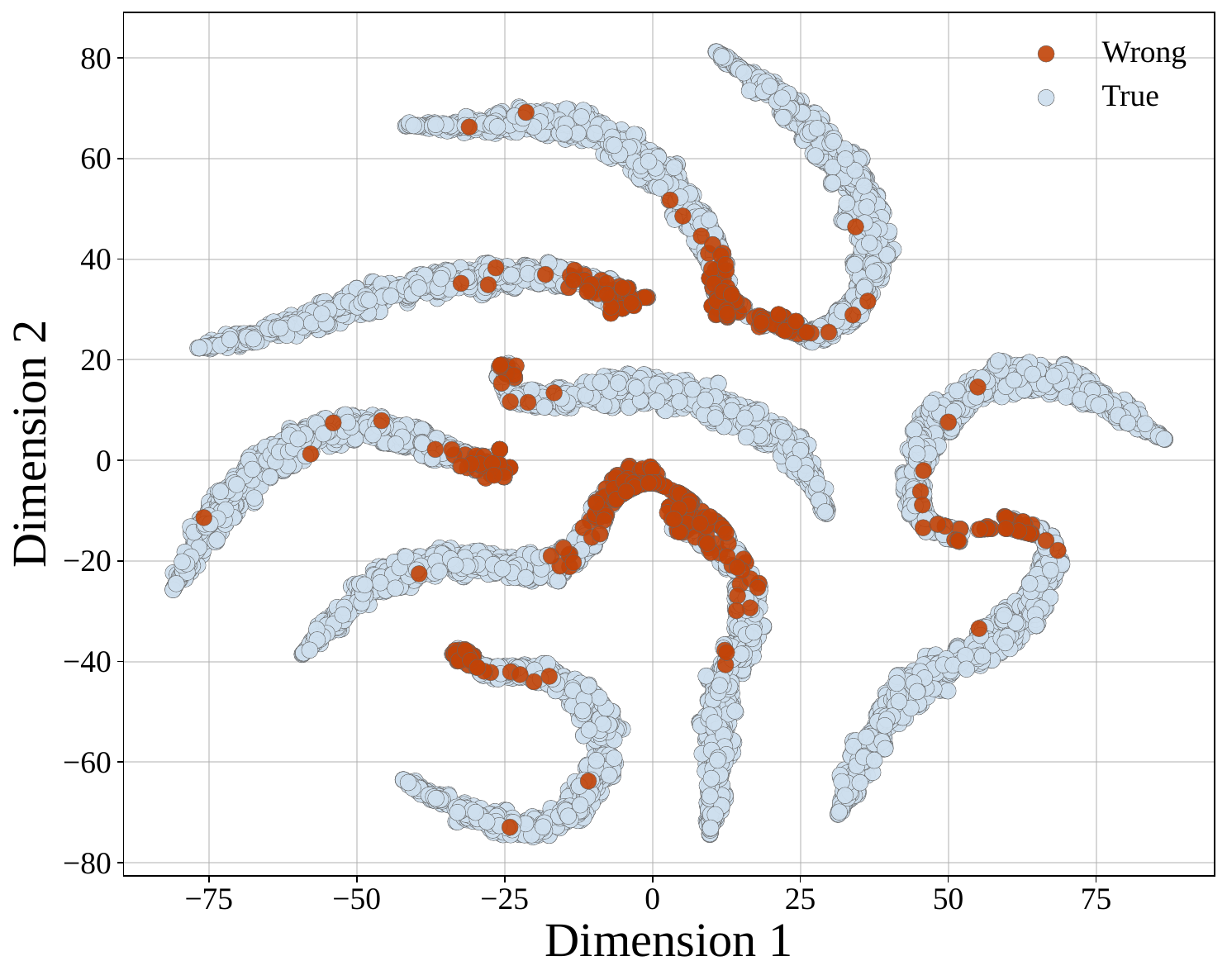}
      \caption{t-SNE of ResNet-18 output vectors on CIFAR10 using the final checkpoint.}
      \label{a:motivation}
    \end{subfigure} &
    \begin{subfigure}[b]{0.47\textwidth}
      \centering
      \includegraphics[width=\linewidth]{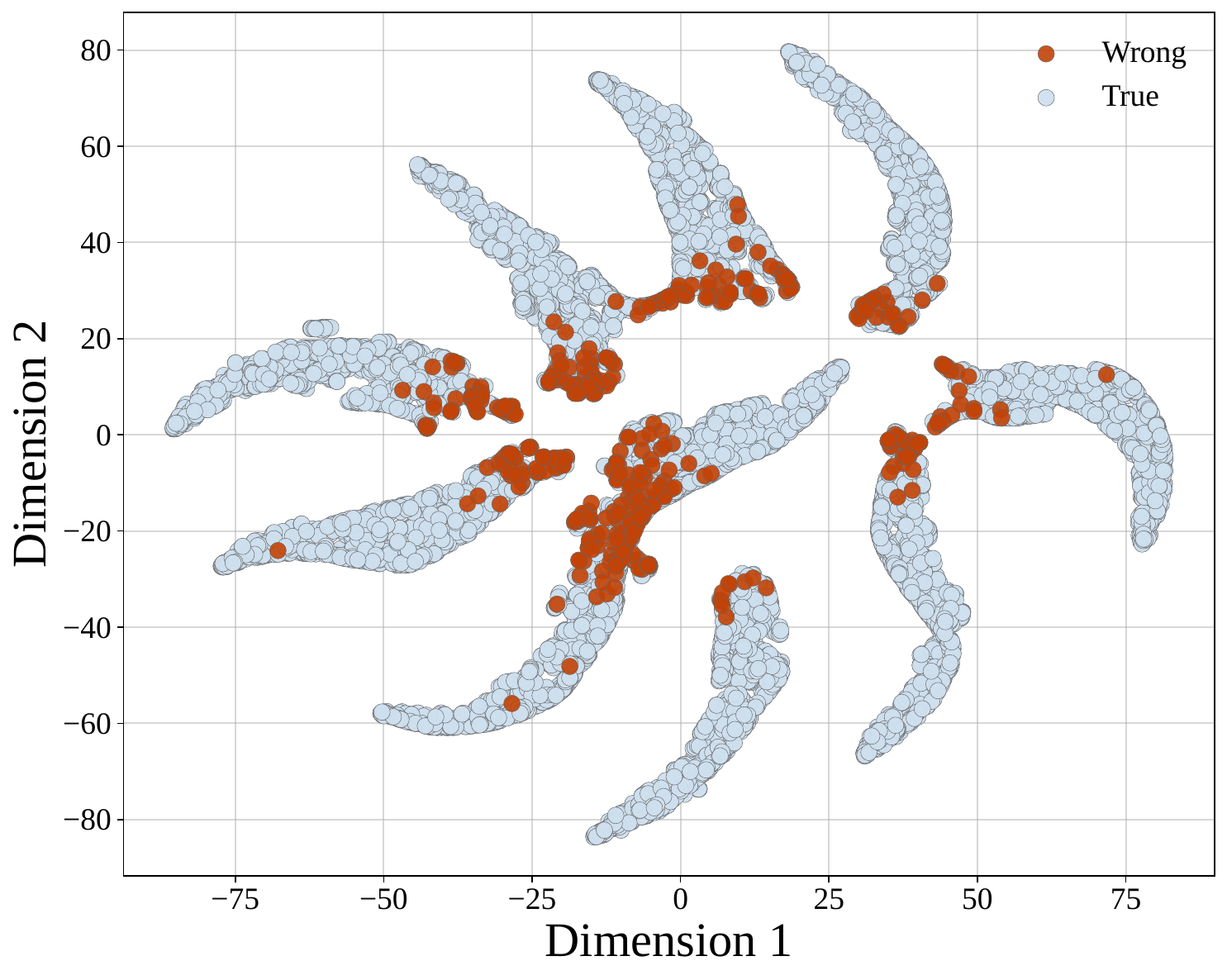}
      \caption{t-SNE using output vectors aggregated over NC-guided checkpoints}
      \label{b:motivation}
    \end{subfigure}
  \end{tabular}
  \caption{Motivation for cross-checkpoint prediction variability on CIFAR10. Red points denote misclassified samples and white points denote correctly classified samples. Compared with a single final checkpoint, aggregating predictions over NC-guided checkpoints improves the separability of failure-prone inputs.}
  \label{motivation}
\end{figure}

In this work, we revisit DNN test prioritization from the perspective of terminal training geometry. Recent studies show that many classifiers exhibit Neural Collapse (NC) in late training. In this phase, class features and classifier weights become highly symmetric, and the decision rule approaches a nearest class center classifier. This suggests that training checkpoints are not arbitrary. They can share geometric properties that help characterize model behavior. Motivated by this insight, we propose an NC-inspired prioritization framework (\textbf{NCIP}). NCIP first selects an NC-guided representative subset of checkpoints using classifier-weight equiangularity. It then uses prediction variability across selected checkpoints to improve ranking reliability.

NCIP replaces single-checkpoint confidence with cross-checkpoint prediction variability. It first selects an NC-guided representative subset of late checkpoints, then measures prediction variability across this subset. Truly easy inputs tend to remain stable across selected checkpoints, while failure-prone inputs are more likely to show prediction changes under checkpoint-induced boundary shifts (see Fig.~\ref{motivation}). Thus, cross-checkpoint prediction variability better reflects boundary proximity and ambiguity, reducing over-prioritization of low-confidence but correct samples.


We evaluate NCIP on multiple datasets and testing settings. Across these scenarios, NCIP achieves strong performance in early fault discovery, achieving higher RAUC than competitive baselines.

In summary, this paper makes the following contributions:
\begin{itemize}
  \item We connect DNN test case prioritization with terminal phase training geometry and show how NC can inform reliable ranking criteria.
  \item We propose NCIP, which selects NC-guided checkpoints using classifier weight equiangularity and prioritizes tests by cross-checkpoint prediction variability.
  \item We conduct extensive experiments demonstrating that NCIP improves early fault discovery and remains effective across fault types, while offering stable and interpretable prioritization signals.
\end{itemize}

\section{Preliminaries}
\subsection{Deep Neural Networks}
DNNs are artificial neural networks characterized by multiple hidden layers, designed to tackle complex pattern recognition tasks by emulating the structure and function of the human brain. They consist of neurons organized hierarchically into input layers, hidden layers, and output layers. Each neuron receives input from the previous layer, computes a weighted sum, applies an activation function, and produces an output for the next layer.

Mathematically, for the $l$-th layer, with input $x^{(l-1)}$, output $x^{(l)}$, weight matrix $W^{(l)}$, bias vector $b^{(l)}$, and activation function $g$, the forward propagation can be represented by Equation \ref{eq:dnn}:

\begin{align}
  \label{eq:dnn}
  x^{(l)} &= g(z^{(l)}) = g(W^{(l)}x^{(l-1)}+b^{(l)})
\end{align}

where $z^{(l)}$ is the result of the linear transformation, and $x^{(l)}$ is the activation value. 

\subsection{Test Case Prioritization}
A large body of work has investigated the testing of DNN-based systems~\cite{pei2017deepxplore, zheng2022neuronfair, harel2020neuron, guo2018dlfuzz, ma2018deepgauge, qiu2022detecting, you2023regression, hu2024test}. Since such systems are developed under a data-driven paradigm~\cite{nielsen2015neural, alabdulmohsin2022revisiting}, a central challenge is to expose model failures effectively under limited testing cost.

Test case prioritization aims to order a candidate test set so that faults are revealed as early as possible within a constrained budget. For a classification model, a fault occurs when the model misclassifies an input. Formally, given a target model $M$, a candidate test set $T$, and a budget $B$ $(B \ll |T|)$, the goal is to rank $T$ and select the top-$B$ subset $T_S$ such that $T_S$ maximizes the number of faults exposed in $M$. Test prioritization has been widely studied in DNN testing as an effective means to accelerate fault discovery under limited time and computation~\cite{feng2020deepgini, bao2023defense, zheng2023certpri, al2022deepabstraction, wang2020dissector, li2024distance, shen2024prioritizing, hwang2023improved, li2023deeprank}.

\subsection{Neural Collapse}
Neural Collapse (NC) refers to a set of geometric regularities that emerge in the terminal phase of training for classifiers, typically after the training error becomes (near) zero while the loss keeps decreasing~\cite{papyan2020prevalence, zhu2021geometric}. Consider a $C$-class model with feature extractor $h(\cdot)\in\mathbb{R}^d$ and linear classifier $W\in\mathbb{R}^{C\times d}$ (rows $\{w_c\}_{c=1}^C$). Let $D_c$ be the training samples of class $c$ and define
\begin{align}
  \mu_c \;=\; \frac{1}{|D_c|}\sum_{x_i\in D_c} h(x_i), \qquad
  \mu_G \;=\; \frac{1}{C}\sum_{c=1}^{C}\mu_c .
\end{align}

NC is commonly summarized by four coupled phenomena~\cite{papyan2020prevalence}:
\begin{itemize}
  \item \textbf{(NC1) Variability collapse.} As training progresses, within-class feature variation becomes negligible, and features concentrate around their corresponding class means.
  \item \textbf{(NC2) Simplex ETF geometry.} The centered class means $\tilde{\mu}_c=\mu_c-\mu_G$ converge to a simplex equiangular tight frame: they have equal norm, form equal angles between any pair, and achieve the maximally pairwise separated configuration under these constraints.
  \item \textbf{(NC3) Self-duality.} The classifier weights and the centered class means converge to each other up to a shared rescaling (and appropriate centering), yielding a highly symmetric decision structure with no systematic preference for confusions between particular class pairs.
  \item \textbf{(NC4) Nearest class center behavior.} The induced decision rule simplifies to choosing the class whose mean is nearest to the input feature in Euclidean distance.
\end{itemize}

Intuitively, NC means that class directions become more evenly spread in late training. Since NC3 links classifier weights to class means, we use classifier-head weights as a practical proxy for this geometry. We then track the spread of pairwise cosine similarities among class-weight vectors. If directions are evenly arranged, cosine values are close and the spread is low, whereas if some directions are unusually close or far apart, cosine values spread out and the spread increases. This gives a simple checkpoint-level indicator of NC regularity.

\section{Methodology}
\subsection{Problem Definition}
We consider unlabeled test case prioritization for a $C$-class classifier $f_\theta$.
Given an unlabeled test set
\begin{equation}
  \label{eq:problem_def}
  X=\{x_i\}_{i=1}^{N}, \qquad x_i\in\mathcal{X},
\end{equation}


Our goal is to find a ranking function $\pi:[N]\to[N]$, yielding the ordered test sequence $\langle x_{\pi(1)},\ldots,x_{\pi(N)}\rangle$. A fault occurs when the model prediction differs from the ground-truth label $y_i$, i.e.,

\begin{equation}
  \label{eq:fault_def}
  \mathbb{I}_i=\mathbb{I}\!\left[f_\theta(x_i)\neq y_i\right],
\end{equation}
where $\mathbb{I}[\cdot]$ is the indicator function.
Since labels are unavailable during prioritization, $\pi$ must be computed without access to $\{y_i\}_{i=1}^{N}$.
The effectiveness of $\pi$ is evaluated by how quickly faults appear in the ranked list under a limited inspection budget $B\ll N$.

\begin{figure}[tb]
  \centering
  \includegraphics[width=0.95\linewidth]{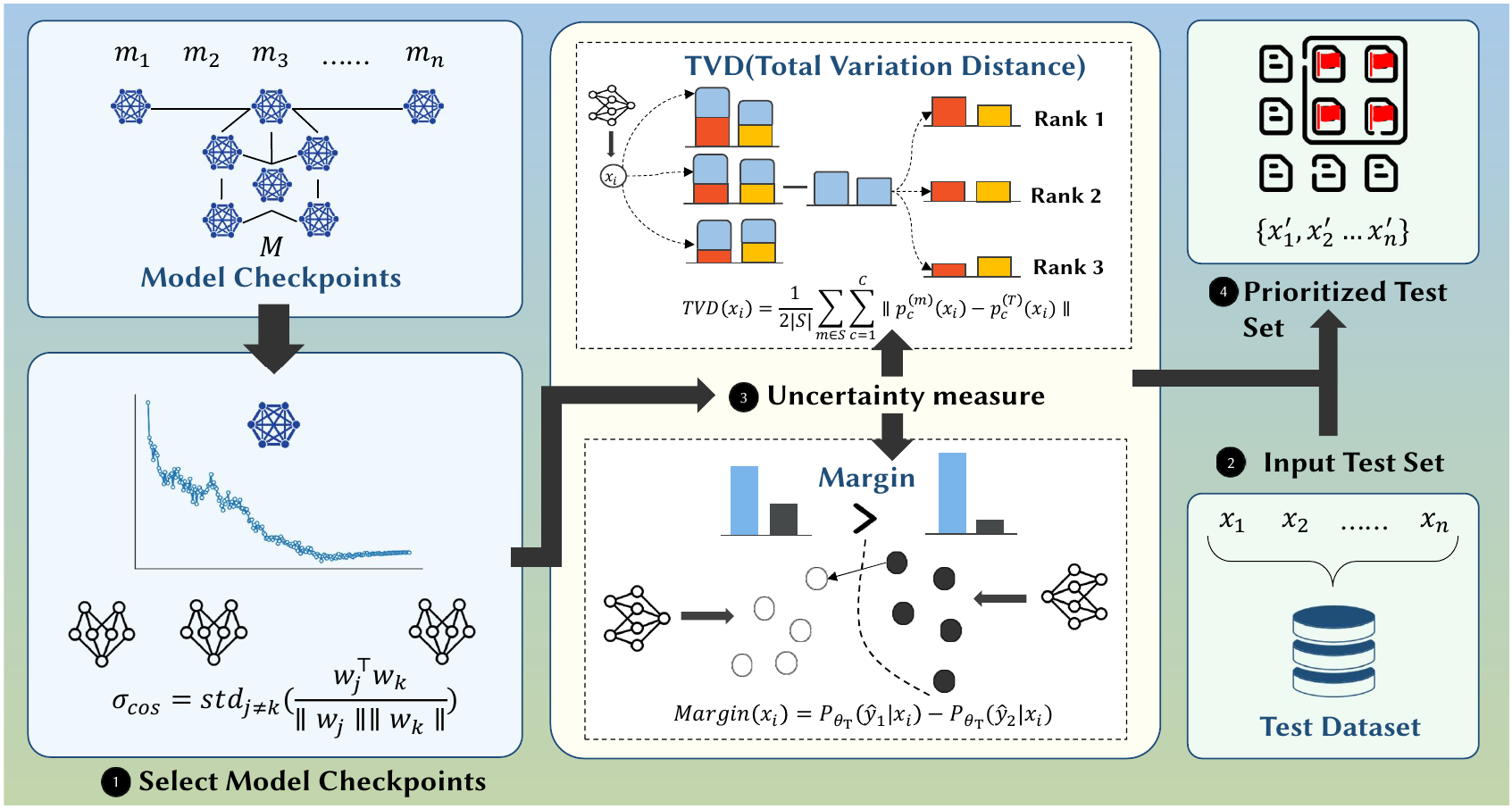}
  \caption{Overview of NCIP. (1) NC checkpoint selection via weight equiangularity to obtain NC-guided model states; (2) cross-checkpoint prediction variability measurement to quantify instability for each test input; and (3) boundary uncertainty integration using the final model margin to refine the prioritization score.}
  \label{fig:overview}
\end{figure}

\subsection{Overview of NCIP Framework}
Our approach, Neural-Collapse-Inspired-Prioritization (\textbf{NCIP}), prioritizes test samples by measuring their inconsistency with NC geometry in the training convergence phase, as shown in Fig.~ \ref{fig:overview}.

Our framework follows a three-stage pipeline. 
First, we perform NC checkpoint selection by extracting training checkpoints and selecting a representative subset based on classifier-weight equiangularity, then use cross-checkpoint prediction variability across this subset for prioritization.
Second, we measure cross-checkpoint prediction variability by quantifying, for each test input, the instability of its predicted probability distribution across the selected checkpoints via Total Variation Distance (TVD), which captures sensitivity to checkpoint-induced decision boundary shifts.
Third, we apply boundary uncertainty integration by incorporating the final checkpoint's prediction margin to emphasize samples close to the decision boundary, where small perturbations are more likely to flip the predicted label. 
The resulting prioritization score ranks inputs that remain unstable across NC-guided checkpoints as higher risk test cases.

\subsection{NC Checkpoint Selection via Weight Equiangularity}



NC theory shows that, in the late training stage, the classifier weight vectors corresponding to different classes form an approximately equiangular configuration. This geometric property provides a robust indicator of whether a model has entered a stable NC regime. Let $W=[w_1,...,w_C]$ denote the classifier weight vectors. 

To evaluate the equiangularity of the classifier weights, we measure the variability of the angles between weight vectors. Specifically, we compute the standard deviation of the pairwise cosine similarities between all distinct pairs of class weight vectors:

\begin{equation}
  \label{eq:NC-aware Checkpoint Selection}
  \sigma_{cos}=\mathrm{std}_{j\neq k}(\frac{w^{\top}_{j}w_k}{\|w_j\|\|w_k\|})
\end{equation}

A smaller value indicates stronger equiangularity and thus closer adherence to NC geometry.

For a set of candidate checkpoints, we compute an NC score for each checkpoint from classifier-head weight geometry using Eq.~\ref{eq:NC-aware Checkpoint Selection}. We first include the final checkpoint in the selected set, then iteratively add the checkpoint whose NC score has the largest minimum distance to the current set. This yields representatives that remain terminally consistent while covering diverse NC states. In our implementation, we use the last \SI{90}{\percent} of available checkpoints as the candidate pool and set the target size to $K=30$ (e.g., if training has 100 total epochs, the candidate pool contains 90 checkpoints, epochs 11--100), as empirically supported in Section~\ref{sec5.4}. After selection, we build and cache a single model incorporating all selected checkpoint weights. In later runs, we directly load this cache instead of rebuilding it, reducing repeated model-loading overhead. Algorithm \ref{alg:nc_select} summarizes the implemented NC checkpoint selection procedure.

\begin{algorithm}
  \caption{NC Checkpoint Selection via Weight Equiangularity}
  \label{alg:nc_select}
  \begin{algorithmic}[1]
    \Inputs
      Checkpoint set $\mathcal{M}=\{\mathcal{M}_1,\dots,\mathcal{M}_m\}$,
      collapse metric $C(\cdot)$,
      target checkpoint count $K$
    \Outputs
      selected checkpoint subset $\mathcal{S}$

    \State Compute $c_j \leftarrow C(\mathcal{M}_j)$ for all $j\in\{1,\dots,m\}$ \RightComment{NC score per checkpoint}

    \State Initialize $T \leftarrow m$
    \State $\mathcal{S}\leftarrow\{T\}$ \RightComment{Keep last checkpoint}

    \While{$|\mathcal{S}| < K$}
      \State For each $u\notin\mathcal{S}$, compute $d(u,\mathcal{S})=\min_{s\in\mathcal{S}}|c_u-c_s|$
      \State $u^* \leftarrow \arg\max_{u\notin\mathcal{S}} d(u,\mathcal{S})$
      \State $\mathcal{S}\leftarrow\mathcal{S}\cup\{u^*\}$
    \EndWhile

    \State Sort $\mathcal{S}$ by epoch index in ascending order
    \State \Return $\mathcal{S}$
  \end{algorithmic}
\end{algorithm}

\subsection{NC-Guided Prediction Variability}
Given the selected NC-guided checkpoints, we quantify prediction instability for each test input. Let $p^{(m)}(x_i)$ denote the predicted class probability distribution of sample $x_i$ under checkpoint $m$, let $p^{(T)}(x_i)$ denote the prediction of the final checkpoint, and let $S$ denote the NC-guided checkpoint subset. We define the Total Variation Distance (TVD) as:

\begin{equation}
\label{eq:TVD}
\mathrm{TVD}(x_i)
=\frac{1}{2|\mathcal{S}|}\sum_{m\in\mathcal{S}}\sum_{c=1}^{C}
\left|p^{(m)}_{c}(x_i)-p^{(T)}_{c}(x_i)\right|.
\end{equation}

\begin{figure}[tb]
  \centering
  \begin{tabular}{cc}
    \begin{subfigure}[b]{0.47\textwidth}
      \centering
      \includegraphics[width=\linewidth]{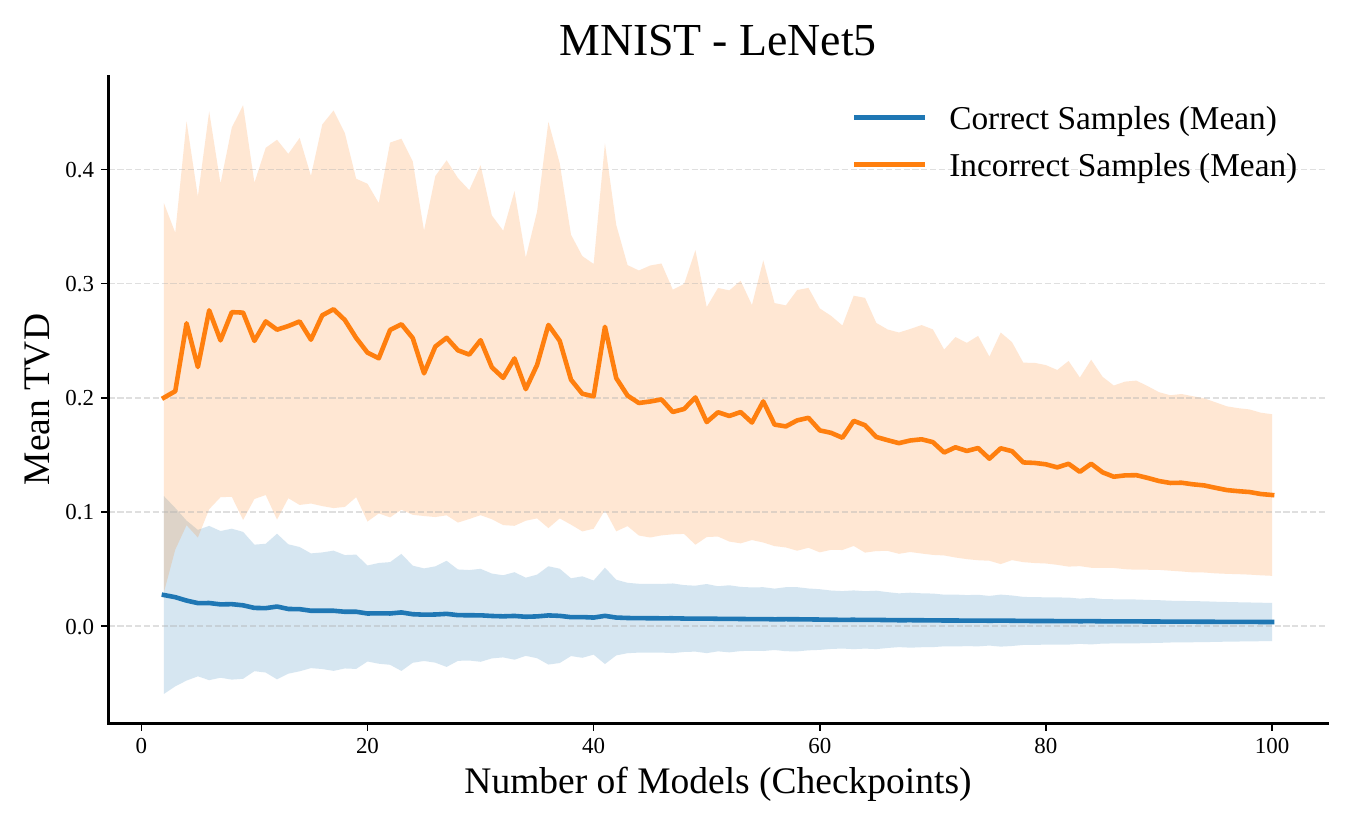}
      \caption{MNIST-LeNet-5: mean TVD (and variability) for correct vs. incorrect samples across checkpoints.}
      \label{a:tvd_correct_vs_incorrect1}
    \end{subfigure} &
    \begin{subfigure}[b]{0.47\textwidth}
      \centering
      \includegraphics[width=\linewidth]{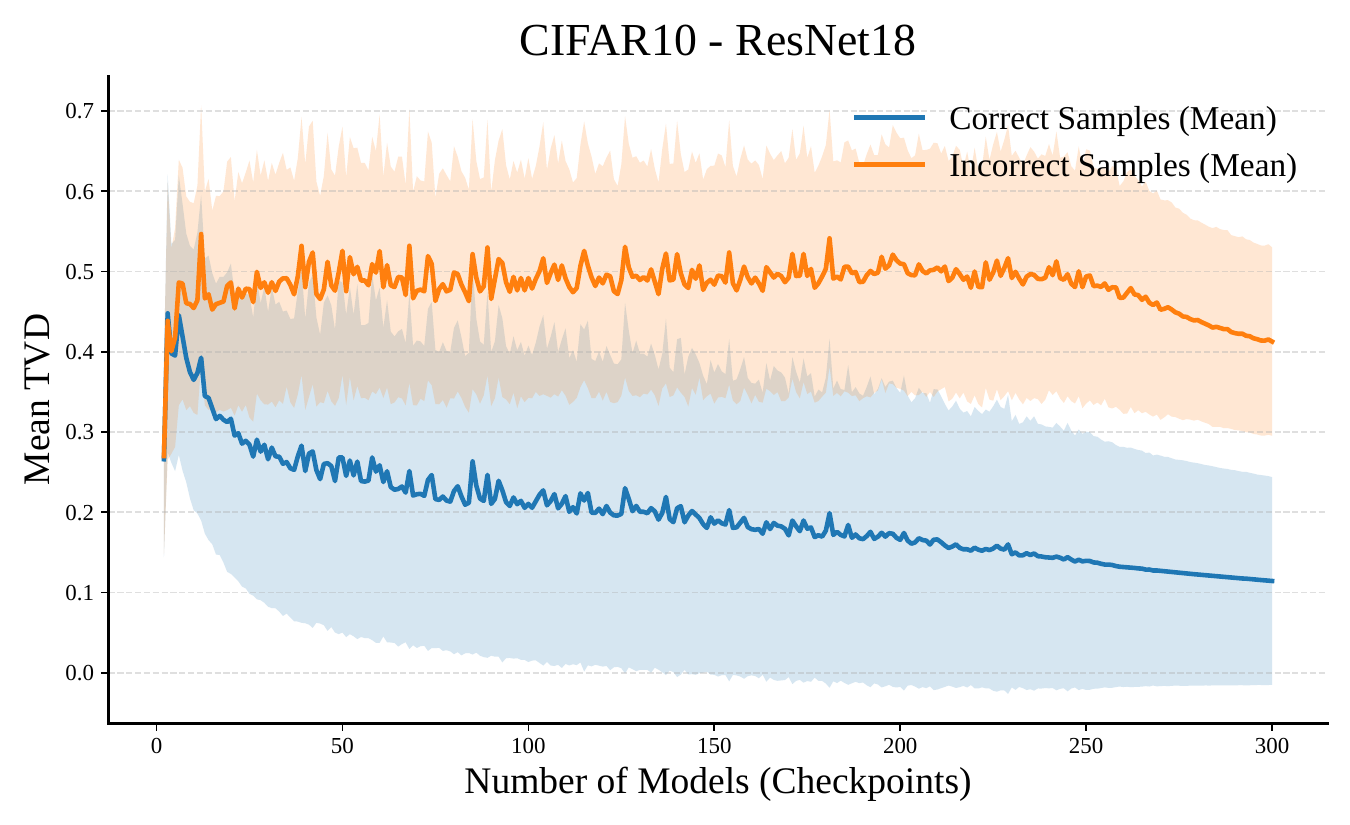}
      \caption{CIFAR10-ResNet-18: mean TVD (and variability) for correct vs. incorrect samples across checkpoints.}
      \label{b:tvd_correct_vs_incorrect2}
    \end{subfigure}
  \end{tabular}
  \caption{Evolution of prediction instability measured by TVD across training checkpoints. In both settings, incorrect samples maintain substantially higher TVD than correct ones as the number of considered checkpoints increases, suggesting that mispredictions can be effectively surfaced by variability-based prioritization.}
  \label{fig:tvd_correct_vs_incorrect}
\end{figure}

For samples consistent with NC geometry, predictions remain stable across NC-guided checkpoints, resulting in small TVD values. In contrast, misclassified or ambiguous samples tend to lie near regions where class attraction competes, causing prediction distributions to fluctuate even among stable models. Fig.~\ref{fig:tvd_correct_vs_incorrect} shows that misclassified samples consistently exhibit larger TVD-based variability than correctly classified ones across late training checkpoints on both CIFAR10-ResNet-18 and MNIST-LeNet-5, supporting TVD as an effective fault-oriented ranking signal. Thus, TVD serves as a direct, sample-level observable of NC instability rather than a heuristic uncertainty measure; we prioritize samples with higher TVD values, which indicate greater instability and a higher risk of misclassification.

While TVD captures global prediction instability, it does not explicitly measure the uncertainty of the target model itself (i.e., the final epoch checkpoint). To complement this, we introduce a boundary uncertainty measure based on the final model's prediction margin.

For each test sample $x_i$, let $p_{(1)}(x_i)$ and $p_{(2)}(x_i)$ denote the largest and second largest predicted probabilities, respectively. The margin is defined as follows:

\begin{equation}
  \label{eq:Margin}
  \mathrm{Margin}(x_i) = p_{(1)}(x_i) - p_{(2)}(x_i)
\end{equation}

A smaller margin implies that the input lies closer to the decision boundary and is therefore more susceptible to prediction flips. This quantity is inexpensive to compute at a single checkpoint and provides a boundary proximity signal complementary to the cross-checkpoint variability captured by TVD.


\subsection{Final Error Prioritization Score}
We define the final prioritization score of NCIP, which prioritizes unlabeled test inputs by combining the global instability signal (TVD) and the local boundary signal (margin), for each test input $x_i$ as

\begin{equation}
  \label{eq:Final_Score}
 \mathrm{NCIP}(x_i) = \mathrm{zscore}(\mathrm{TVD}(x_i)) + \mathrm{zscore}(1 - \mathrm{Margin}(x_i))
\end{equation}
Here, zscore(·) denotes the standardization process. The test samples are then ranked in descending order of $NCIP(x_i)$, with higher-ranked samples prioritized for testing.

This formulation requires no training dataset. By integrating training dynamics and decision boundary information, it provides an effective and robust solution for unlabeled error prioritization. Algorithm \ref{alg:nc_tvd_fusion} outlines our algorithm.

\begin{table}[tb] 
  \centering 
  \caption{Dataset and DNN models}
  \label{tab:datasets_models}
  \fontsize{7}{8.5}\selectfont
  \renewcommand{\arraystretch}{1.0}
  \setlength{\tabcolsep}{2.5pt}
  \begin{tabular}{M{1.5cm}M{3.5cm}M{0.9cm}|M{1.5cm}M{1.0cm}}
    \toprule 
    Dataset & Description & Epochs & DNN Model & Acc \\
    \midrule

    MNIST  &
    28$\times$28 handwritten digits &
    100 &
    LeNet-1 \quad\quad\quad\quad\quad LeNet-5 &
    0.9887 \quad\quad\quad\quad 0.9918 \\ 

    \midrule
    Fashion-MNIST &
    28$\times$28 grayscale images &
    200 &
    LeNet-1 \quad\quad\quad\quad\quad LeNet-5 &
    0.8917 \quad\quad\quad\quad 0.9021 \\ 

    \midrule
    CIFAR10 &
    32$\times$32 colored images &
    300 &
    ResNet-18 \quad\quad\quad\quad VGG-11 &
    0.9554 \quad\quad\quad\quad 0.8936 \\  

    \midrule
    CIFAR100 &
    32$\times$32 colored images with 100 classes &
    500 &
    ResNet-50 \quad\quad\quad\quad DenseNet-121 &
    0.7793 \quad\quad\quad\quad 0.7888 \\  

    \midrule
    TinyImageNet &
    High-resolution images with 200 classes &
    500 &
    ResNet-152 \quad\quad\quad\quad DenseNet-201 &
    0.5484 \quad\quad\quad\quad 0.5814 \\

    \midrule
    IMDB &
    Movie reviews for sentiment analysis (binary) &
    100 &
    TextCNN \quad\quad\quad\quad Transformer &
    0.8030 \quad\quad\quad\quad 0.7546 \\

    \midrule
    AGNEWS &
    News topic classification (4 classes) &
    100 &
    TextCNN \quad\quad\quad\quad Transformer &
    0.9154 \quad\quad\quad\quad 0.9030 \\

    \bottomrule 
  \end{tabular} 
\end{table}

\begin{algorithm}
  \caption{NC-guided Error Prioritization via NCIP}
  \label{alg:nc_tvd_fusion}
  \begin{algorithmic}[1]
    \Inputs
      Selected checkpoints $\mathcal{S}$,
      test set $T=\{x_i\}_{i=1}^{N}$
    \Outputs
      Ranking indices $\pi$,
      TVD scores $\mathbf{u}^{tvd}\in\mathbb{R}^N$,
      margins $\mathbf{m}\in\mathbb{R}^N$

    \State Initialize $\mathbf{u}^{tvd} \leftarrow \mathbf{0}$ \RightComment{TVD per sample}
    \State Initialize $\mathbf{m} \leftarrow \mathbf{0}$ \RightComment{Margin per sample}

    \State Choose reference checkpoint $r \leftarrow \max(\mathcal{S})$ \RightComment{Use last selected epoch as reference}

    \For{each sample $x_i \in T$}
      \State $p^{(r)} \leftarrow \text{softmax}(\mathcal{M}_r(x_i))$ \RightComment{Reference distribution}

      \For{each checkpoint $j \in \mathcal{S}, j \neq r$}
        \State $p^{(j)} \leftarrow \text{softmax}(\mathcal{M}_j(x_i))$
        \State $d_{ij} \leftarrow \frac{1}{2}\|p^{(j)} - p^{(r)}\|_1$ \RightComment{Total variation distance}
        \State $\mathbf{u}^{tvd}_i \leftarrow \mathbf{u}^{tvd}_i + d_{ij}$
      \EndFor

      \State $(p_{(1)}, p_{(2)}) \leftarrow \text{Top2}(p^{(r)})$ \RightComment{Largest and second largest probs}
      \State $\mathbf{m}_i \leftarrow p_{(1)} - p_{(2)}$ \RightComment{Prediction margin}
    \EndFor

    \State $\mathbf{s} \leftarrow \mathrm{zscore}(\mathbf{u}^{tvd}) + \mathrm{zscore}(\mathbf{1} - \mathbf{m})$ \RightComment{Final risk score}
    \State $\pi \leftarrow \text{argsort}(\mathbf{s}, \text{descending})$

    \State \Return $\pi, \mathbf{u}^{tvd}, \mathbf{m}$
  \end{algorithmic}
\end{algorithm}

\section{Experiment Design}

\subsection{Datasets and Models}
To ensure a fair comparison, we used datasets and models consistent with previous studies \cite{bao2023defense, li2024distance, shen2024prioritizing, wang2025sets}, as detailed in Table \ref{tab:datasets_models}. We employed seven well-known image and text classification datasets: MNIST \cite{lecun1998gradient}, Fashion-MNIST \cite{xiao2017fashion}, CIFAR10 \cite{krizhevsky2009learning}, CIFAR100 \cite{krizhevsky2009learning}, TinyImageNet \cite{le2015tiny}, IMDB \cite{maas2011learning} and AGNEWS \cite{zhang2015character}. The experiments involved widely used DNN architectures: LeNet-1 \cite{lecun1998gradient} for MNIST, LeNet-5 \cite{lecun1998gradient} for  Fashion-MNIST, ResNet-18 \cite{he2016deep} and VGG-11 \cite{simonyan2014very} for CIFAR10, ResNet-50 \cite{he2016deep} and DenseNet-121 \cite{huang2017densely} for CIFAR100, and ResNet-152 \cite{he2016deep} and DenseNet-201 \cite{huang2017densely} for TinyImageNet. For the text classification datasets, we adopted both CNN-based~\cite{kim2014convolutional} and Transformer-based~\cite{vaswani2017attention} models.

For LLM generative-task evaluation, we additionally use the WikiText dataset~\cite{merity2016pointer} with two fine-tuned causal language models, GPT-2~\cite{radford2019language} and OPT-125m~\cite{zhang2022opt}.

\subsection{Adversarial Data}
\label{sec:Adversarial Data}
Adversarial data consists of deliberately altered inputs with subtle perturbations that lead deep learning models to make incorrect predictions despite these changes being nearly imperceptible to humans. To assess the effectiveness of our proposed DNN testing method in detecting errors with adversarial samples, we included four common attack methods: FGSM \cite{goodfellow2014explaining}, PGD \cite{madry2017towards}, CW \cite{carlini2017towards}, and BIM \cite{kurakin2018adversarial}, implemented using the torchattack library \cite{kim2020torchattacks} with default hyperparameters.

Table~\ref{tab:adv_attack_config} summarizes the attack settings used in adversarial evaluation. All attacks are run with standard default settings and in untargeted mode. For each dataset-model pair, we generate adversarial examples for FGSM, PGD, BIM, and CW separately, and then combine them into one adversarial test set. We report RAUC-500 and RAUC-ALL in Section~\ref{sec:RQ1} on this combined set.

\begin{table}[tb]
  \centering
  \caption{Adversarial attack configurations}
  \label{tab:adv_attack_config}
  \fontsize{7}{8.5}\selectfont
  \renewcommand{\arraystretch}{1.3}
  \setlength{\tabcolsep}{2.5pt}
  \begin{tabular}{lcccc}
    \toprule
    \textbf{Attack} & \textbf{Norm} & \textbf{Perturbation Budget} & \textbf{Targeted} & \textbf{Other Settings} \\
    \midrule
    FGSM & $L_{\infty}$ & $\epsilon=8/255$ & Untargeted & Single-step \\
    PGD & $L_{\infty}$ & $\epsilon=8/255$ & Untargeted & $\alpha=2/255$, steps=10, random start \\
    BIM & $L_{\infty}$ & $\epsilon=8/255$ & Untargeted & $\alpha=2/255$, steps=10 \\
    CW & $L_{2}$ & Optimization-based $L_2$ objective & Untargeted & $c=1$, $\kappa=0$, steps=50, lr=0.01 \\
    \bottomrule
  \end{tabular}
\end{table}

\subsection{Baseline Methods}
We selected several classic methods, including uncertainty-based methods: DeepGini \cite{feng2020deepgini}, Entropy \cite{byun2019input}, PCS \cite{zhang2020towards}, and MSP \cite{weiss2022simple}, as well as SA-based methods: DSA and LSA \cite{kim2019guiding}, implemented via a third-party library.

To differentiate our method from approaches that estimate disagreement from a single checkpoint, we include Dropout \cite{hu2023aries} and EffiMAP \cite{wei2022predictive} as baselines. Dropout estimates uncertainty from repeated stochastic forward passes and uses it as a proxy for the distance to the decision boundary. EffiMAP perturbs both model and input, measures the resulting prediction changes, and prioritizes samples with larger changes.

In addition, we incorporate several representative baselines for comparison. NNS \cite{hwang2023improved} considers not only the uncertainty of a DNN on a given test input but also the model's uncertainty over its neighboring samples. TDPR \cite{shen2024prioritizing} constructs a learning trajectory for each test input to characterize the evolving learning dynamics of DNNs across training, from which informative features are extracted and fed into a learning-to-rank framework to derive a prioritized ordering. SETS \cite{wang2025sets} focuses on high uncertainty test inputs to first shrink the candidate pool and then applies an efficient greedy strategy to further reduce the number of fitness evaluations.

\subsection{Evaluation Metrics}
\label{sec:Evaluation Metrics}
We assessed the prioritization performance across four key dimensions: effectiveness, robustness, diversity, and guidance.

\textbf{Effectiveness and Robustness.} To assess the prioritization ranking's effectiveness and robustness, we employed the Ratio of the Area Under Curve (RAUC) \cite{zheng2023certpri, shen2024prioritizing} as a performance metric for evaluating test sample selection methods. Here, robustness refers explicitly to the method's effectiveness when the input samples are replaced by adversarial data. For classification tasks, the RAUC is mathematically defined as follows:

\begin{align}
	RAUC=\frac{\sum_{i=1}^{N} n_{i}}{N \times N^{\prime}+\frac{N^{\prime}-N^{\prime 2}}{2}}, \text { where } n_{i}=\left\{\begin{array}{l}
		n_{i-1}+1, c(x_{i}) \text{ is incorrect} \\
		n_{i-1}, \text{ otherwise}
		\end{array}\right.
\end{align}

Here, $N$ is the total number of test inputs, and $N'$ represents the number of misclassified samples. The term $c(x_i)$ indicates the model's classification prediction for input $x_i$. This metric prioritizes test inputs that are more likely to be misclassified, with $\sum_{i=1}^{N} n_{i}$ denoting the sum of indices of misclassified samples in the ranked results.

Given the high labeling cost of test inputs and resource constraints, it is essential to identify erroneous samples quickly. To address this, we introduce RAUC-n (RAUC for the top $n$ prioritized test inputs) to measure prioritization effectiveness under limited budget scenarios. Specifically, we evaluate RAUC-500 and RAUC-ALL.

\textbf{Diversity.} While the above metrics evaluate the effectiveness of test methods in identifying erroneous samples, we also employ the $Fault\_Type$ metric~\cite{gao2022adaptive, zhang2025efficient} to assess the diversity of detected faults. If misclassified samples are ranked highly but correspond to the same fault type, the prioritization method exhibits limited fault diversity and is therefore less effective. If a test sample $x$ is misclassified as another label, $Fault\_Type$ is defined as follows:

\begin{align}
	Fault\_Type(x)=(y\rightarrow f(x)), \text{where }  f(x)\neq y
\end{align}

Here, $y$ denotes the actual label of the input $x$, and $f(x)$ indicates the predicted label resulting from the misclassification. 




\textbf{Guidance.} Under each budget, we retrain the DNN with corrected labels from the top-ranked uncertain samples and report test-accuracy gain, defined as retrained accuracy minus original accuracy.

\subsection{Research Questions}
We investigate the effectiveness and practicality of our approach through the following five research questions (RQs):

\textbf{RQ1: Effectiveness on Clean and Adversarial Data.} How effectively does our method prioritize error-prone inputs on both clean and adversarial samples?

\textbf{RQ2: Fault Diversity Evaluation.} Can our method detect a wider variety of error types compared to other approaches?

\textbf{RQ3: Effectiveness on DNN Enhancement.} Does our method outperform existing approaches in terms of guiding data selection for DNN enhancement?

\textbf{RQ4: Ablation Study and Parameter Impact.} How does varying the number of selected checkpoints affect the performance of our method?

\textbf{RQ5: Efficiency.} How efficient is our method in prioritizing test inputs?

\section{Experiment Results}
We present our experimental results and analyze the outcomes. We implemented our method using PyTorch V1.12.1 in Python. The experiments were conducted on an Ubuntu 22.04 system equipped with 8 x NVIDIA GeForce RTX 4090 GPUs and \SI{512}{GB} DDR5 RAM. We repeated the process five times for experiments and took the average.

\begin{table}[tb]
  \centering
  \caption{RAUC-ALL on Clean Data}
  \label{RAUC-ALL on Clean Data}
  \fontsize{7}{8.5}\selectfont
  \renewcommand{\arraystretch}{1.3}
  \setlength{\tabcolsep}{2.5pt}

  \begin{tabular}{c!{\vrule}cccccccccc!{\vrule}cccc}
      \toprule 
      \multirow{2}{*}{\textbf{Methods}} 
      & \multicolumn{2}{c}{\textbf{MNIST}} 
      & \multicolumn{2}{c}{\textbf{FMNIST}} 
      & \multicolumn{2}{c}{\textbf{CIFAR10}} 
      & \multicolumn{2}{c}{\textbf{CIFAR100}} 
      & \multicolumn{2}{c!{\vrule}}{\textbf{TinyImageNet}} 
      & \multicolumn{2}{c}{\textbf{IMDB}} 
      & \multicolumn{2}{c}{\textbf{AGNEWS}} \\

      & LeNet-1 & LeNet-5 
      & LeNet-1 & LeNet-5 
      & RN-18 & VGG-11 
      & RN-50 & DN-121 
      & RN-152 & DN-201
      & CNN & Trans.
      & CNN & Trans. \\

      \midrule 

      RS 
      & 0.513 & 0.575 & 0.547 & 0.515 & 0.528 & 0.519 & 0.560 & 0.561 & 0.650 & 0.635
      & 0.554 & 0.574 & 0.507 & 0.541 \\

      \midrule 

      LSA 
      & 0.900 & 0.851 & 0.845 & 0.774 & 0.787 & 0.771 & 0.864 & 0.860 & 0.865 & 0.839
      & 0.720 & 0.607 & 0.804 & 0.700 \\

      DSA 
      & 0.979 & 0.982 & 0.885 & 0.893 & 0.943 & 0.907 & 0.891 & 0.887 & 0.893 & 0.878
      & 0.755 & 0.728 & 0.838 & 0.830 \\

      \midrule 

      Entropy 
      & 0.983 & 0.985 & 0.896 & 0.898 & 0.941 & {\cellcolor[rgb]{0.753,0.753,0.753}}\textbf{0.910} & 0.884 & 0.880 & 0.897 & 0.874
      & 0.785 & 0.725 & 0.859 & 0.827 \\

      DeepGini 
      & 0.983 & 0.985 & 0.897 & 0.898 & 0.942 & 0.909 & 0.888 & 0.885 & 0.898 & 0.874
      & 0.785 & 0.725 & 0.862 & 0.828 \\

      MSP 
      & 0.983 & 0.985 & 0.897 & 0.898 & 0.942 & 0.909 & 0.889 & 0.886 & 0.898 & 0.873
      & 0.785 & 0.725 & 0.863 & 0.828 \\

      PCS 
      & 0.982 & 0.985 & 0.895 & 0.897 & 0.942 & 0.909 & 0.890 & 0.887 & 0.894 & 0.871
      & 0.785 & 0.725 & 0.864 & 0.828 \\

      \midrule

      Dropout
      & 0.785 & 0.941 & 0.581 & 0.848 & 0.855 & 0.720 & 0.487 & 0.510 & 0.578 & 0.693
      & 0.690 & 0.705 & 0.702 & 0.830 \\

      EffiMAP
      & 0.857 & 0.753 & 0.855 & 0.861 & 0.885 & 0.866 & 0.870 & 0.862 & 0.884 & 0.870
      & -- & -- & -- & -- \\

      \midrule 

      NNS 
      & 0.981 & 0.984 & 0.899 & 0.910 & 0.943 & 0.909 & 0.893 & 0.892 & 0.896 & 0.879
      & 0.785 & 0.724 & 0.867 & 0.829 \\

      TDPR 
      & 0.981 & 0.879 & 0.869 & 0.904 & 0.922 & 0.871 & 0.883 & 0.877 & 0.892 & 0.898
      & 0.798 & 0.748 & 0.862 & 0.862 \\

      SETS 
      & 0.954 & 0.952 & 0.889 & 0.891 & 0.919 & 0.901 & 0.883 & 0.880 & 0.895 & 0.870
      & 0.785 & 0.725 & 0.845 & 0.816 \\

      \midrule 

      NCIP 
      & {\cellcolor[rgb]{0.753,0.753,0.753}}\textbf{0.986} & {\cellcolor[rgb]{0.753,0.753,0.753}}\textbf{0.986} 
      & {\cellcolor[rgb]{0.753,0.753,0.753}}\textbf{0.904} & {\cellcolor[rgb]{0.753,0.753,0.753}}\textbf{0.931} 
      & {\cellcolor[rgb]{0.753,0.753,0.753}}\textbf{0.952} & 0.908 
      & {\cellcolor[rgb]{0.753,0.753,0.753}}\textbf{0.902} & {\cellcolor[rgb]{0.753,0.753,0.753}}\textbf{0.904} 
      & {\cellcolor[rgb]{0.753,0.753,0.753}}\textbf{0.908} & {\cellcolor[rgb]{0.753,0.753,0.753}}\textbf{0.906}
      & {\cellcolor[rgb]{0.753,0.753,0.753}}\textbf{0.814} & {\cellcolor[rgb]{0.753,0.753,0.753}}\textbf{0.844} 
      & {\cellcolor[rgb]{0.753,0.753,0.753}}\textbf{0.879} & {\cellcolor[rgb]{0.753,0.753,0.753}}\textbf{0.898} \\

      \bottomrule 
  \end{tabular}
\end{table}

\begin{table}[tb]
  \centering
  \caption{RAUC-500 on Clean Data}
  \label{RAUC-500 on Clean Data}
  \footnotesize
  \fontsize{7}{8.5}\selectfont
  \renewcommand{\arraystretch}{1.3}
  \setlength{\tabcolsep}{2.5pt}
  \begin{tabular}{c!{\vrule}cccccccccc!{\vrule}cccc}
      \toprule 
      \multirow{2}{*}{\textbf{Methods}} 
      & \multicolumn{2}{c}{\textbf{MNIST}} 
      & \multicolumn{2}{c}{\textbf{FMNIST}} 
      & \multicolumn{2}{c}{\textbf{CIFAR10}} 
      & \multicolumn{2}{c}{\textbf{CIFAR100}} 
      & \multicolumn{2}{c!{\vrule}}{\textbf{TinyImageNet}} 
      & \multicolumn{2}{c}{\textbf{IMDB}} 
      & \multicolumn{2}{c}{\textbf{AGNEWS}} \\
      & LeNet-1 & LeNet-5 & LeNet-1 & LeNet-5 & RN-18 & VGG-11 & RN-50 & DN-121 & RN-152 & DN-201
      & CNN & Trans. & CNN & Trans.\\
      \midrule 
      RS & 0.022 & 0.056 & 0.119 & 0.067 & 0.063 & 0.092 & 0.202 & 0.188 & 0.445 & 0.460
      & 0.169 & 0.252 & 0.097 & 0.074 \\
      \midrule
      LSA & 0.571 & 0.462 & 0.503 & 0.537 & 0.502 & 0.617 & 0.706 & 0.700 & 0.670 & 0.858
      & 0.445 & 0.506 & 0.465 & 0.465 \\
      DSA & 0.677 & 0.685 & 0.561 & 0.558 & 0.540 & 0.604 & 0.750 & 0.708 & 0.853 & 0.834
      & 0.492 & 0.511 & 0.465 & 0.461 \\
      \midrule
      Entropy & 0.707 & 0.718 & 0.566 & 0.563 & 0.530 & 0.622 & 0.765 & 0.774 & 0.886 & 0.903
      & 0.496 & 0.522 & 0.445 & 0.482 \\
      DeepGini & 0.706 & 0.720 & 0.564 & 0.561 & 0.541 & \cellcolor[rgb]{0.753,0.753,0.753}\textbf{0.630} & 0.781 & 0.793 & 0.888 & 0.896
      & 0.496 & 0.522 & 0.468 & 0.489 \\
      MSP & 0.705 & 0.720 & 0.572 & 0.557 & 0.545 & 0.626 & 0.789 & 0.801 & 0.901 & 0.883
      & 0.496 & 0.522 & 0.481 & 0.487 \\
      PCS & 0.700 & 0.720 & 0.542 & 0.548 & 0.541 & 0.606 & 0.758 & 0.735 & 0.905 & 0.820
      & 0.496 & 0.522 & 0.486 & 0.484 \\
      \midrule
      Dropout & 0.175 & 0.514 & 0.065 & 0.533 & 0.206 & 0.196 & 0.135 & 0.280 & 0.280 & 0.434
      & 0.326 & 0.525 & 0.158 & 0.462 \\
      EffiMAP & 0.394 & 0.163 & 0.460 & 0.472 & 0.432 & 0.528 & 0.695 & 0.685 & 0.860 & 0.846
      & -- & -- & -- & -- \\
      \midrule
      NNS & 0.713 & 0.709 & 0.574 & 0.585 & 0.536 & 0.607 & \cellcolor[rgb]{0.753,0.753,0.753}\textbf{0.801} & 0.796 & 0.878 & 0.902
      & 0.497 & 0.511 & 0.493 & 0.490 \\
      TDPR & 0.722 & 0.677 & 0.409 & \cellcolor[rgb]{0.753,0.753,0.753}\textbf{0.655} & 0.461 & 0.600 & 0.745 & 0.798 & \cellcolor[rgb]{0.753,0.753,0.753}\textbf{0.916} & 0.886
      & 0.519 & 0.550 & 0.432 & \cellcolor[rgb]{0.753,0.753,0.753}\textbf{0.623} \\
      SETS & 0.497 & 0.444 & 0.477 & 0.454 & 0.367 & 0.503 & 0.612 & 0.628 & 0.819 & 0.787
      & 0.473 & 0.523 & 0.355 & 0.353 \\
      \midrule
      NCIP & \cellcolor[rgb]{0.753,0.753,0.753}\textbf{0.757}
              & \cellcolor[rgb]{0.753,0.753,0.753}\textbf{0.739}
              & \cellcolor[rgb]{0.753,0.753,0.753}\textbf{0.611}
              & 0.638
              & \cellcolor[rgb]{0.753,0.753,0.753}\textbf{0.563}
              & 0.621
              & 0.796
              & \cellcolor[rgb]{0.753,0.753,0.753}\textbf{0.839}
              & 0.904
              & \cellcolor[rgb]{0.753,0.753,0.753}\textbf{0.917}
              & \cellcolor[rgb]{0.753,0.753,0.753}\textbf{0.704}
              & \cellcolor[rgb]{0.753,0.753,0.753}\textbf{0.584}
              & \cellcolor[rgb]{0.753,0.753,0.753}\textbf{0.525}
              & 0.610 \\
      \bottomrule 
  \end{tabular}
\end{table}

\subsection{Effectiveness on Clean and Adversarial Data (RQ1)}
\label{sec:RQ1}
\textbf{Effectiveness on Clean Data.} To evaluate the effectiveness of NCIP in identifying varied errors, we compare it against state-of-the-art baselines across all subject datasets and models. We employ two complementary metrics: RAUC-ALL to assess the overall quality of the test prioritization ranking, and RAUC-500 to measure practical efficiency under a limited labeling budget.

Table~\ref{RAUC-ALL on Clean Data} reports the RAUC-ALL results for all competing methods. As shown, NCIP consistently outperforms the strongest baselines across most dataset-model pairs. Compared with widely adopted uncertainty-based metrics such as DeepGini and Entropy, NCIP achieves superior ranking performance. For example, on CIFAR10–ResNet-18, NCIP attains a RAUC-ALL score of 0.952, outperforming both DeepGini and Entropy. This result suggests that leveraging the geometric properties induced by NC, especially the equiangular structure of classifier weights, provides a more robust indicator of error proneness than softmax confidence alone. NCIP also demonstrates advantages over coverage-guided techniques such as DSA and LSA. On FMNIST–LeNet-5, NCIP achieves a RAUC-ALL score of 0.931, substantially exceeding LSA and remaining competitive with DSA.

\textbf{Efficiency Under Limited Budget.} In practical testing scenarios, testers typically face a constrained budget for inspecting and labeling test inputs. Table~\ref{RAUC-500 on Clean Data} presents the RAUC-500 results, reflecting the errors identified within the top 500 prioritized samples. Although NCIP does not achieve the absolute best performance on every individual dataset-model pair, it consistently excels at the early discovery of faults. For challenging configurations such as CIFAR100–DenseNet-121, NCIP achieves a RAUC-500 score of 0.839, surpassing the strongest baseline MSP, which attains 0.801. This improvement can be attributed to NCIP's explicit exploitation of NC geometry to prioritize geometrically distinct samples, thereby prioritizing fault-revealing test cases under a limited budget.

\begin{table}[t]
  \centering
  \caption{Wilcoxon Test Results on RAUC-ALL for NCIP vs. Each Baseline (One-sided, Holm-corrected).}
  \label{tab:wilcoxon_short}
  \scriptsize
  \setlength{\tabcolsep}{3pt}
  \begin{tabular}{lccc|lccc}
    \hline
    Baseline & $p_{\text{raw}}$ & $p_{\text{Holm}}$ & Effect Size & Baseline & $p_{\text{raw}}$ & $p_{\text{Holm}}$ & Effect Size \\
    \hline
    DeepGini & $1.22\times10^{-4}$ & $7.32\times10^{-4}$ & $0.981$ & NNS & $1.22\times10^{-4}$ & $7.32\times10^{-4}$ & $0.981$ \\
    Dropout & $6.10\times10^{-5}$ & $7.32\times10^{-4}$ & $1.000$ & PCS & $1.22\times10^{-4}$ & $7.32\times10^{-4}$ & $0.981$ \\
    DSA & $6.10\times10^{-5}$ & $7.32\times10^{-4}$ & $1.000$ & Random & $6.10\times10^{-5}$ & $7.32\times10^{-4}$ & $1.000$ \\
    EffiMAP & $9.77\times10^{-4}$ & $9.77\times10^{-4}$ & $1.000$ & SETS & $6.10\times10^{-5}$ & $7.32\times10^{-4}$ & $1.000$ \\
    Entropy & $1.22\times10^{-4}$ & $7.32\times10^{-4}$ & $0.981$ & TDPR & $6.10\times10^{-5}$ & $7.32\times10^{-4}$ & $1.000$ \\
    LSA & $6.10\times10^{-5}$ & $7.32\times10^{-4}$ & $1.000$ & MSP & $1.22\times10^{-4}$ & $7.32\times10^{-4}$ & $0.981$ \\
    \hline
  \end{tabular}
\end{table}

\begin{figure}[tb]
  \centering
  \begin{subfigure}[b]{0.664\textwidth}
    \centering
    \includegraphics[width=1.0\linewidth]{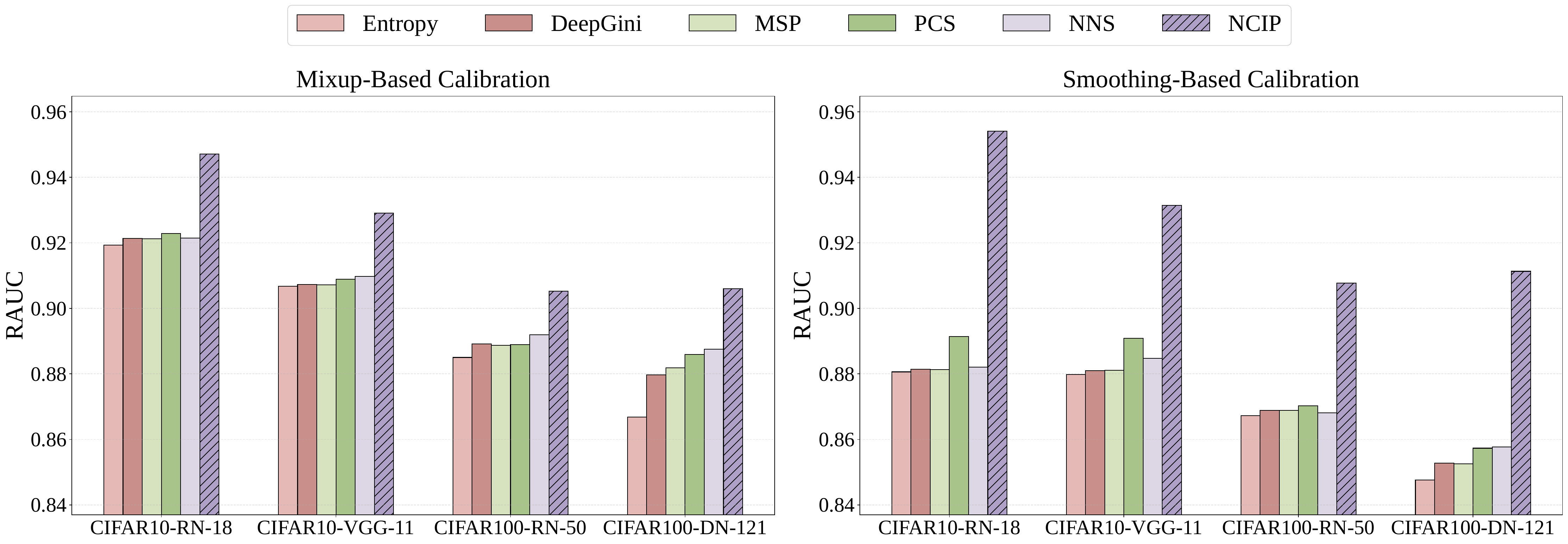}
    \caption{}
    \label{fig:mix}
  \end{subfigure} 
  \hfill
  \begin{subfigure}[b]{0.31\textwidth}
    \centering
    \includegraphics[width=1.0\linewidth]{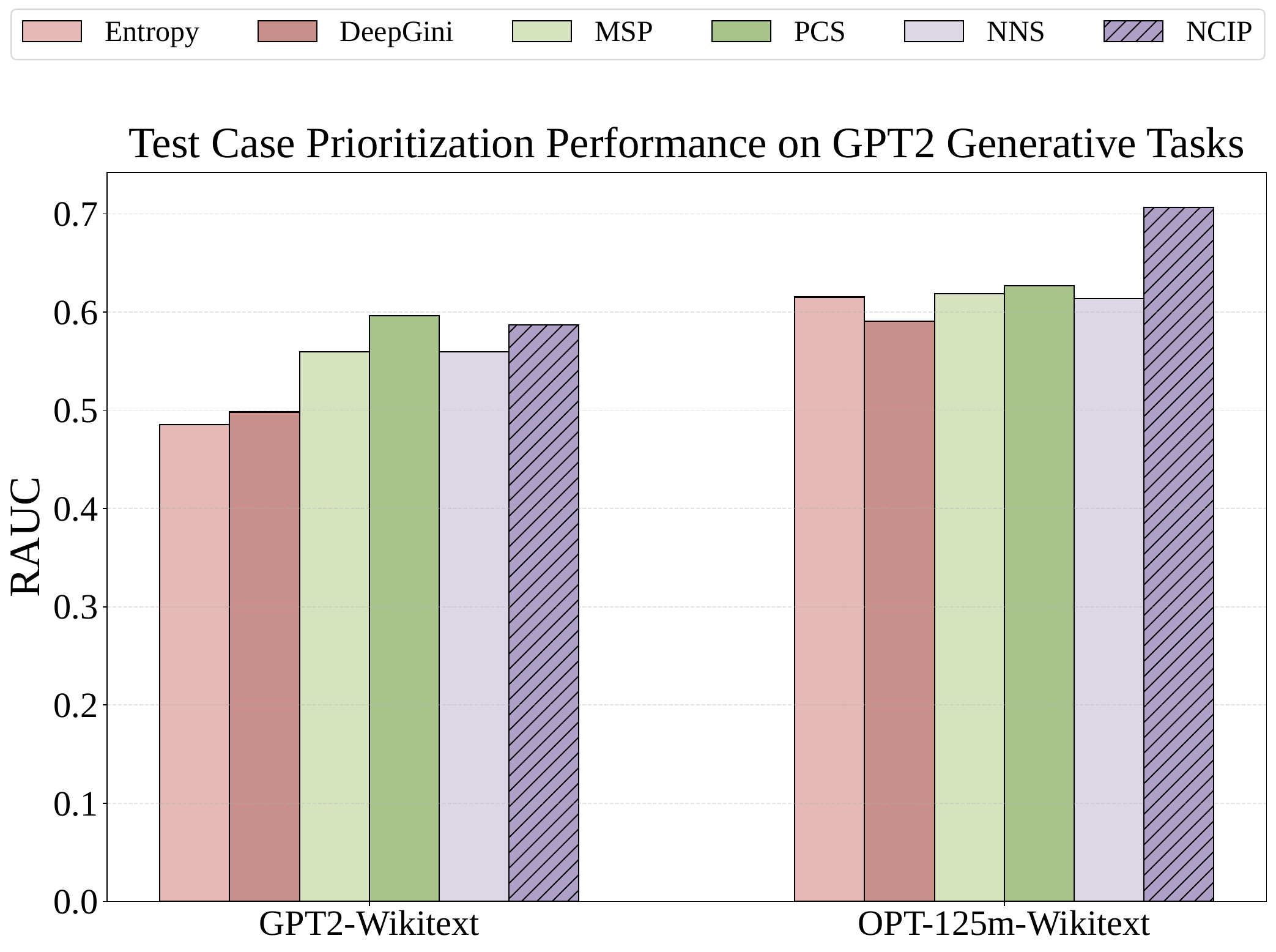}
    \caption{}
    \label{fig:llm}
  \end{subfigure} 
  \caption{(a) Test case prioritization under model calibration. (b) NCIP performance on LLM generative tasks using only fine-tuning checkpoints}
\end{figure}

\textbf{Generalization Across Modalities.} It is worth noting that NCIP maintains high performance across diverse modalities. On the text classification task AGNEWS-Transformer, NCIP achieves the highest RAUC-ALL of 0.898, demonstrating its generalizability beyond image datasets. This is evident in AGNEWS-TextCNN, where NCIP achieves a RAUC-500 score of 0.525, significantly outperforming DeepGini with a score of 0.468 and Entropy with a score of 0.445, thereby demonstrating its ability to select an earlier set of error-triggering inputs under the same testing budget.

To further validate statistical significance, we applied Wilcoxon signed-rank tests~\cite{macfarland2016wilcoxon} at $\alpha=0.05$. We compared NCIP with each baseline on paired RAUC-ALL results (mostly $n=14$ pairs; $n=10$ for EffiMAP). To handle multiple comparisons, we applied Holm correction and judged significance using adjusted $p$-values. Exact $p$-values and effect sizes (rank-biserial correlation, RBC) are reported in Table~\ref{tab:wilcoxon_short}.

After Holm correction, RAUC-ALL shows significant improvements for all baselines ($p_{\text{Holm}}\le 9.77\times10^{-4}$) with large positive effects (RBC $=0.981$--$1.000$). This result provides consistent statistical evidence that NCIP improves overall ranking quality across dataset-model pairs.


To examine the impact of model calibration on test prioritization methods, we additionally evaluate two common calibration strategies, mixup~\cite{zhang2017mixup} and label smoothing~\cite{szegedy2016rethinking} (Fig.~\ref{fig:mix}). NCIP remains competitive after calibration, suggesting that the improvement is not only due to confidence rescaling at a single checkpoint, but also due to the cross-checkpoint instability signal.

We further explore extended usage scenarios of NCIP. NCIP requires access to multiple checkpoints, so its primary deployment setting is model development and model-serving pipelines where training or fine-tuning trajectories are retained. In the LLM scenario, we do not assume access to pretraining checkpoints. Instead, we use only fine-tuning checkpoints, as shown in Fig.~\ref{fig:llm}, and treat low BLEU-2 outputs as failures. Under this setting, NCIP remains competitive on WikiText, achieving 0.587 with GPT-2 (best: 0.596) and 0.706 with OPT-125m (best). These results indicate that NCIP can be applied to generative tasks when fine-tuning checkpoints are available.

\begin{table}[tb]
  \centering
  \caption{RAUC results on adversarial data. Highlighted values indicate the best results, and underlined values indicate the second-best results.}
  \label{tab:rauc_adv_combined}
  \footnotesize
  \fontsize{7}{8.5}\selectfont
  \renewcommand{\arraystretch}{1.3}
  \setlength{\tabcolsep}{2.5pt}

  \begin{tabular}{c!{\vrule}ccccccc!{\vrule}ccccccc}
      \toprule
      \multirow{3}{*}{\textbf{Methods}} 
      & \multicolumn{7}{c!{\vrule}}{\textbf{RAUC-500}} 
      & \multicolumn{7}{c}{\textbf{RAUC-ALL}} \\
      
      & \multicolumn{2}{c}{CIFAR10}
      & \multicolumn{2}{c}{CIFAR100}
      & \multicolumn{2}{c}{TinyImageNet}
      & Avg.
      & \multicolumn{2}{c}{CIFAR10}
      & \multicolumn{2}{c}{CIFAR100}
      & \multicolumn{2}{c}{TinyImageNet}
      & Avg. \\

      & RN-18 & VGG-11 & RN-50 & DN-121 & RN-152 & DN-201 & 
      & RN-18 & VGG-11 & RN-50 & DN-121 & RN-152 & DN-201 & \\

      \midrule
      RS 
      & 0.686 & 0.666 & 0.816 & 0.786 & 0.933 & 0.883 & 0.795
      & 0.748 & 0.747 & 0.834 & 0.844 & 0.915 & 0.914 & 0.834 \\

      \midrule
      LSA 
      & 0.932 & 0.927 & 0.927 & 0.906 & 0.881 & 0.965 & 0.923
      & 0.829 & 0.837 & 0.900 & 0.902 & 0.928 & 0.922 & 0.886 \\

      DSA 
      & 0.926 & \underline{0.972} & \cellcolor[rgb]{0.753,0.753,0.753}\textbf{0.994} & 0.973 & 0.941 & 0.970 & 0.963
      & 0.843 & 0.875 & 0.905 & 0.908 & 0.936 & 0.921 & 0.898 \\

      \midrule
      Entropy 
      & \underline{0.944} & 0.932 & 0.969 & 0.974 & 0.985 & 0.974 & 0.963
      & 0.836 & 0.874 & 0.904 & 0.907 & 0.939 & 0.912 & 0.895 \\

      DeepGini 
      & 0.942 & 0.931 & 0.969 & 0.979 & 0.986 & 0.977 & \underline{0.964}
      & 0.836 & 0.873 & 0.904 & 0.907 & 0.938 & 0.912 & 0.895 \\

      MSP 
      & 0.936 & 0.927 & 0.973 & \cellcolor[rgb]{0.753,0.753,0.753}\textbf{0.985} & \cellcolor[rgb]{0.753,0.753,0.753}\textbf{0.988} & 0.976 & \underline{0.964}
      & 0.836 & 0.874 & 0.904 & 0.907 & 0.937 & 0.911 & 0.895 \\

      PCS 
      & \cellcolor[rgb]{0.753,0.753,0.753}\textbf{0.945} & \cellcolor[rgb]{0.753,0.753,0.753}\textbf{0.974}
      & 0.970 & 0.973 & 0.877 & 0.924 & 0.944
      & 0.835 & 0.872 & 0.903 & 0.905 & 0.935 & 0.911 & 0.893 \\

      \midrule
      Dropout
      & 0.712 & 0.768 & 0.389 & 0.543 & 0.801 & 0.892 & 0.684
      & 0.841 & 0.834 & 0.780 & 0.789 & 0.904 & 0.930 & 0.846 \\

      EffiMAP
      & 0.879 & 0.755 & 0.957 & 0.955 & 0.972 & 0.956 & 0.912
      & 0.852 & 0.858 & 0.898 & 0.908 & 0.937 & 0.936 & 0.898 \\

      \midrule
      NNS 
      & 0.887 & 0.892 & 0.949 & 0.956 & 0.960 & 0.963 & 0.935
      & 0.822 & 0.801 & 0.887 & 0.889 & 0.928 & 0.905 & 0.872 \\

      TDPR 
      & 0.936 & 0.928 & \underline{0.974} & \underline{0.982}
      & \cellcolor[rgb]{0.753,0.753,0.753}\textbf{0.988} & \cellcolor[rgb]{0.753,0.753,0.753}\textbf{0.988}
      & \cellcolor[rgb]{0.753,0.753,0.753}\textbf{0.966}
      & \cellcolor[rgb]{0.753,0.753,0.753}\textbf{0.937} & \cellcolor[rgb]{0.753,0.753,0.753}\textbf{0.899}
      & \cellcolor[rgb]{0.753,0.753,0.753}\textbf{0.953} & \cellcolor[rgb]{0.753,0.753,0.753}\textbf{0.961}
      & \cellcolor[rgb]{0.753,0.753,0.753}\textbf{0.959} & \cellcolor[rgb]{0.753,0.753,0.753}\textbf{0.964}
      & \cellcolor[rgb]{0.753,0.753,0.753}\textbf{0.946} \\

      SETS 
      & 0.913 & 0.893 & 0.946 & 0.956 & 0.979 & 0.955 & 0.940
      & 0.836 & 0.874 & 0.904 & 0.907 & 0.937 & 0.911 & 0.895 \\

      \midrule
      NCIP 
      & \underline{0.944} & 0.939 & 0.965 & 0.963 & 0.973 & \cellcolor[rgb]{0.753,0.753,0.753}\textbf{0.988} & 0.962
      & \underline{0.926} & \cellcolor[rgb]{0.753,0.753,0.753}\textbf{0.899} & \underline{0.939} & \underline{0.945} & \underline{0.954} & \underline{0.958} & \underline{0.937} \\

      \bottomrule
  \end{tabular}
\end{table}

\textbf{Effectiveness on Adversarial Data.} To evaluate the robustness of NCIP under adversarial attacks, we conducted experiments on adversarial datasets. The results are summarized in Table~\ref{tab:rauc_adv_combined}.

NCIP remains strong under adversarial settings, with average scores reaching 0.962 (RAUC-500) and 0.937 (RAUC-ALL). Although TDPR achieves the best overall average in this table (0.966/0.946), NCIP is better than the remaining baselines across most dataset-model pairs. For example, on CIFAR10-ResNet-18, NCIP achieves 0.926 RAUC-ALL, while entropy-based baselines are around 0.836.


\begin{tcolorbox}
    \noindent \textbf{Answer to RQ1: NCIP consistently outperforms most baselines on both clean and adversarial data. It improves RAUC by \SIrange[range-phrase=--]{1.5}{16.6}{\percent} and increases RAUC-500 by \SIrange[range-phrase=--]{4.9}{20.6}{\percent}, enabling earlier fault discovery. Under adversarial settings, NCIP remains robust with second-best average scores (RAUC-500/RAUC-ALL: 0.962/0.937), and performs well under model calibration and on LLM generative tasks.}
\end{tcolorbox}

\begin{figure}[tb]
  \centering
  \includegraphics[width=0.85\linewidth]{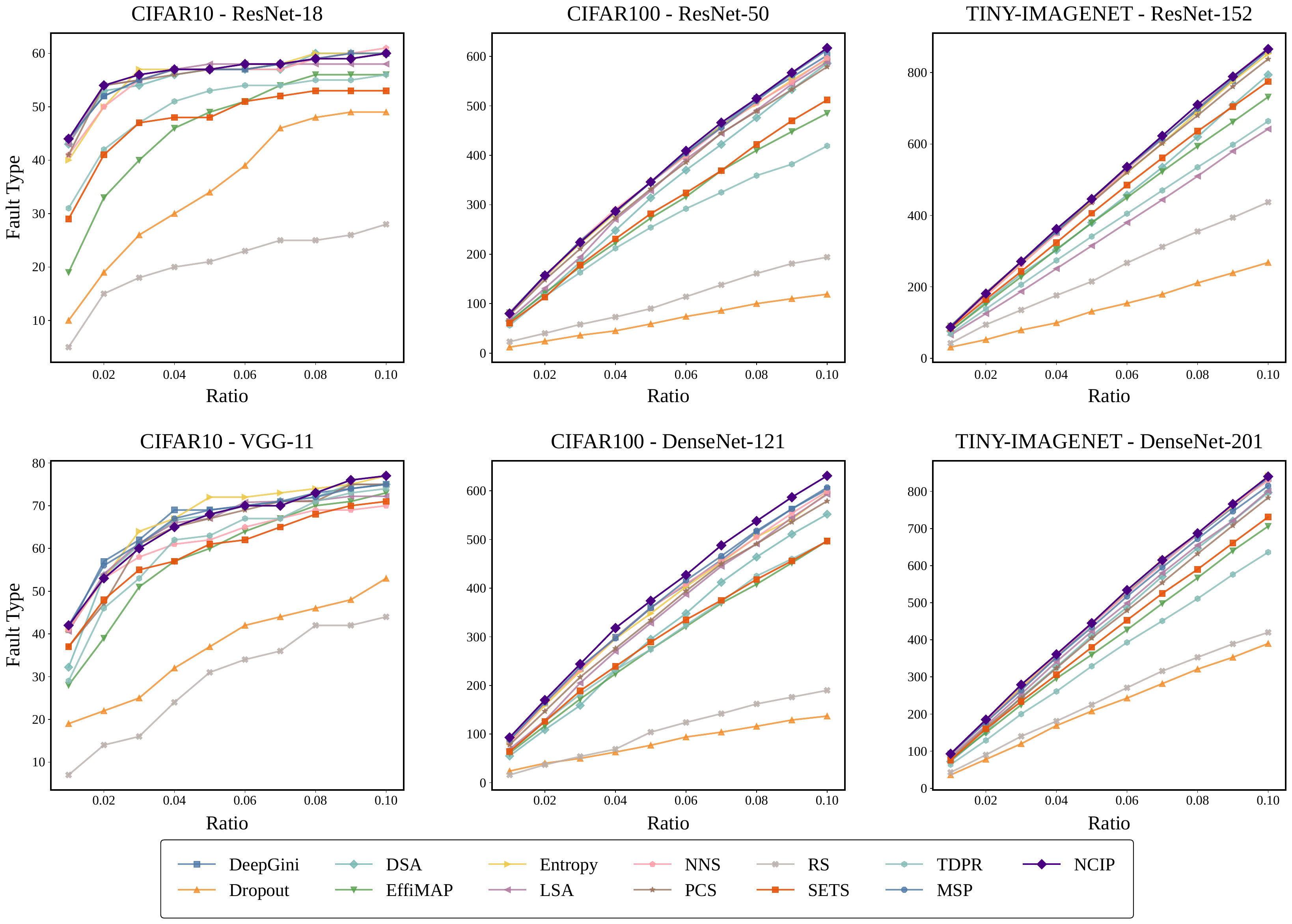}
  \caption{Fault Diversity on Clean Data}
  \label{fig:Fault Diversity on Clean Data}
\end{figure}

\subsection{Fault Diversity Evaluation (RQ2)}
To quantify the diversity of errors detected by different DNN testing methods under varying budgets, we evaluated the number of error types identified within budgets of \SIrange{1}{10}{\percent} of the total clean test samples. As shown in Fig.~\ref{fig:Fault Diversity on Clean Data}, the horizontal axis denotes the test case selection budget as a percentage of the total samples, while the vertical axis represents the Fault Diversity score.

Fig.~\ref{fig:Fault Diversity on Clean Data} illustrates that NCIP achieves greater fault diversity than baseline methods across various dataset scales.

Overall, NCIP outperforms competing approaches in identifying a wider range of fault types for most dataset-model combinations. For example, on CIFAR100 with DenseNet-121, under a stringent budget of \SI{5}{\percent}, NCIP identifies 374 distinct failures, exceeding the performance of the strongest baseline, DeepGini, which detects 360 failures. This result highlights NCIP's effectiveness in prioritizing inputs that expose diverse failure modes at an early stage.
\begin{tcolorbox}
    \noindent \textbf{Answer to RQ2: NCIP enhances fault diversity by \SIrange[range-phrase=--]{17.5}{37.6}{\percent} over baselines, ensuring comprehensive defect discovery for safety-critical systems.}
\end{tcolorbox}

\subsection{Effectiveness on DNN Enhancement (RQ3)}
\begin{figure}[tb]
  \centering
  \includegraphics[width=0.95\linewidth]{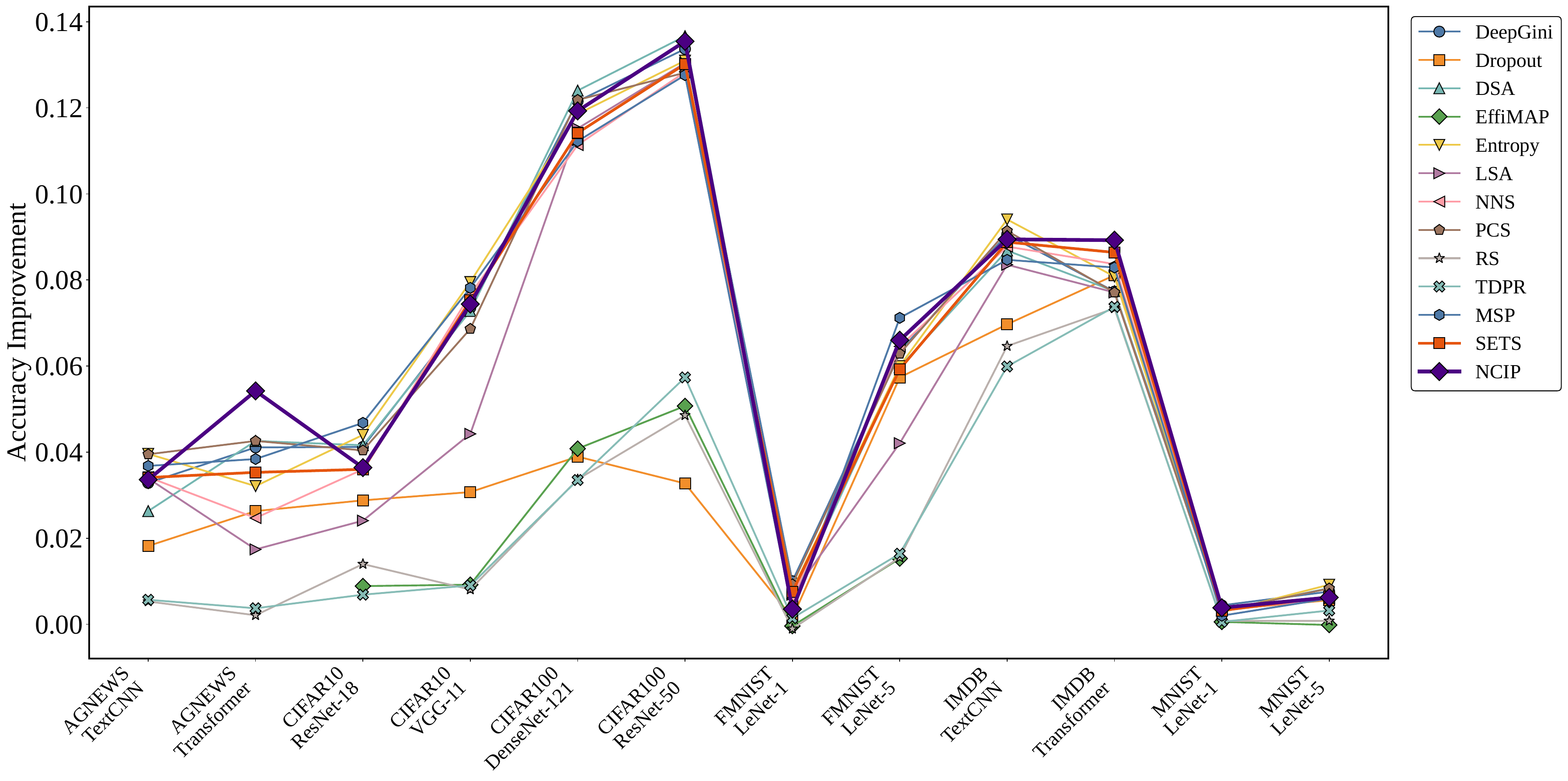}
  \caption{Accuracy improvement after retraining with the top \SI{20}{\percent} selected test cases (averaged over five runs).}
  \label{ACC Improvement on Clean Data}
\end{figure}

To evaluate the effectiveness of the selected test cases for model improvement, we conducted retraining experiments. Specifically, we fine-tune the original models using the top \SI{20}{\percent} of samples prioritized by each method and measure the resulting accuracy improvement on a held-out test set. The results are summarized in Fig.~\ref{ACC Improvement on Clean Data}.

The results indicate that retraining with samples selected by NCIP consistently leads to accuracy gains compared with other methods. NCIP achieves average accuracy improvements of \SIrange[range-phrase=--]{1.52}{5.94}{\percent}. These improvements suggest that samples with high NCIP scores correspond to challenging and informative instances that are insufficiently learned by the original models.


\begin{tcolorbox}
    \noindent \textbf{Answer to RQ3: NCIP provides informative retraining data, achieving \SIrange[range-phrase=--]{1.52}{5.94}{\percent} greater accuracy than baselines, enabling effective model improvement under limited budgets.}
\end{tcolorbox}

\subsection{Ablation Study and Parameter Impact (RQ4)}
\label{sec5.4}
To quantify the contribution of each component in NCIP, we conducted an ablation study on the TVD score, the NC checkpoint selection via weight equiangularity, and the prediction margin. We compare the full NCIP variant (All) with three reduced versions: TVD only (TVD), TVD with NC checkpoint selection (TVD+NC), and TVD with margin (TVD+Margin). Results on fault diversity under a \SI{5}{\percent} budget and RAUC-ALL are reported in Table~\ref{tab:ablation_diverse_5} and Table~\ref{tab:ablation_rauc_all}, respectively.

Incorporating the margin term is particularly beneficial for improving fault diversity, especially on challenging classification tasks. On CIFAR100 with DenseNet-121, adding the margin (TVD+Margin) increases the number of distinct failures identified at \SI{5}{\percent} budget from 260 to 301, indicating that margin information helps prioritize ambiguous samples near the decision boundary.

The TVD metric alone delivers reasonable performance across all datasets, lending support to our hypothesis that prediction variability across checkpoints can serve as a useful indicator of input hardness. For example, on CIFAR10 with ResNet-18, TVD achieves a RAUC-ALL score of 0.952.

\begin{table}[tb]
  \centering
  \caption{Ablation study on fault diversity at \SI{5}{\percent} budget (clean data).}
  \label{tab:ablation_diverse_5}
  \footnotesize
  \fontsize{7}{8.5}\selectfont
  \renewcommand{\arraystretch}{1.3}
  \setlength{\tabcolsep}{2.5pt}
  \begin{tabular}{c!{\vrule}cccccccccc}
    \toprule
    \multirow{2}{*}{\textbf{Variants}} 
    & \multicolumn{2}{c}{\textbf{MNIST}} 
    & \multicolumn{2}{c}{\textbf{FMNIST}} 
    & \multicolumn{2}{c}{\textbf{CIFAR10}} 
    & \multicolumn{2}{c}{\textbf{CIFAR100}}
    & \multicolumn{2}{c}{\textbf{TinyImageNet}} \\
      & LeNet-1 & LeNet-5 & LeNet-1 & LeNet-5 & RN-18 & VGG-11 & RN-50 & DN-121 & RN-152 & DN-201 \\
    \midrule
    TVD & 51 & {\cellcolor[rgb]{0.753,0.753,0.753}}\textbf{42} & 43 & 41 & 56 & 66 & 234 & 260 & 366 & 400 \\
    TVD+NC & 51 & 41 & 46 & {\cellcolor[rgb]{0.753,0.753,0.753}}\textbf{50} & {\cellcolor[rgb]{0.753,0.753,0.753}}\textbf{57} & {\cellcolor[rgb]{0.753,0.753,0.753}}\textbf{71} & 321 & 350 & 425 & 443 \\
    TVD+Margin & 51 & {\cellcolor[rgb]{0.753,0.753,0.753}}\textbf{42} & 42 & 45 & 56 & 66 & 267 & 301 & 392 & 418 \\
    NCIP (Full) & {\cellcolor[rgb]{0.753,0.753,0.753}}\textbf{52} & 41 & {\cellcolor[rgb]{0.753,0.753,0.753}}\textbf{48} & {\cellcolor[rgb]{0.753,0.753,0.753}}\textbf{50} & {\cellcolor[rgb]{0.753,0.753,0.753}}\textbf{57} & 68 & {\cellcolor[rgb]{0.753,0.753,0.753}}\textbf{346} & {\cellcolor[rgb]{0.753,0.753,0.753}}\textbf{374} & {\cellcolor[rgb]{0.753,0.753,0.753}}\textbf{446} & {\cellcolor[rgb]{0.753,0.753,0.753}}\textbf{445} \\
    \bottomrule
  \end{tabular}
\end{table}

\begin{table}[tb]
  \centering
  \caption{Ablation study on RAUC-ALL (clean data).}
  \label{tab:ablation_rauc_all}
  \fontsize{7}{8.5}\selectfont
  \renewcommand{\arraystretch}{1.3}
  \setlength{\tabcolsep}{1.8pt}
  \begin{tabular}{c!{\vrule}ccccccccccccccc}
    \toprule
    \multirow{2}{*}{\textbf{Variants}} 
    & \multicolumn{2}{c}{\textbf{MNIST}} 
    & \multicolumn{2}{c}{\textbf{FMNIST}} 
    & \multicolumn{2}{c}{\textbf{CIFAR10}} 
    & \multicolumn{2}{c}{\textbf{CIFAR100}} 
    & \multicolumn{2}{c}{\textbf{IMDB}} 
    & \multicolumn{2}{c}{\textbf{AGNEWS}} 
    & \multicolumn{2}{c}{\textbf{TinyImageNet}} 
    & \multirow{2}{*}{\textbf{Avg.}} \\
    & LeNet-1 & LeNet-5 & LeNet-1 & LeNet-5 & RN-18 & VGG-11 & RN-50 & DN-121 & CNN & Trans. & CNN & Trans. & RN-152 & DN-201 & \\
    \midrule
    TVD
    & 0.983 & 0.985
    & 0.891 & 0.901
    & 0.944 & 0.907
    & 0.875 & 0.882
    & 0.784 & 0.715
    & 0.864 & 0.892
    & 0.884 & 0.884
    & 0.885 \\
    TVD+NC
    & 0.983 & 0.986
    & 0.880 & 0.929
    & 0.942 & 0.904
    & 0.890 & 0.894
    & 0.798 & {\cellcolor[rgb]{0.753,0.753,0.753}}\textbf{0.851}
    & {\cellcolor[rgb]{0.753,0.753,0.753}}\textbf{0.880} & 0.891
    & 0.907 & {\cellcolor[rgb]{0.753,0.753,0.753}}\textbf{0.912}
    & 0.903 \\
    TVD+Margin
    & 0.983 & 0.985
    & 0.896 & 0.901
    & 0.944 & {\cellcolor[rgb]{0.753,0.753,0.753}}\textbf{0.909}
    & 0.887 & 0.890
    & 0.786 & 0.720
    & 0.866 & 0.889
    & 0.893 & 0.881
    & 0.888 \\
    NCIP (Full)
    & {\cellcolor[rgb]{0.753,0.753,0.753}}\textbf{0.986} & {\cellcolor[rgb]{0.753,0.753,0.753}}\textbf{0.987}
    & {\cellcolor[rgb]{0.753,0.753,0.753}}\textbf{0.904} & {\cellcolor[rgb]{0.753,0.753,0.753}}\textbf{0.931}
    & {\cellcolor[rgb]{0.753,0.753,0.753}}\textbf{0.952} & 0.908
    & {\cellcolor[rgb]{0.753,0.753,0.753}}\textbf{0.902} & {\cellcolor[rgb]{0.753,0.753,0.753}}\textbf{0.904}
    & {\cellcolor[rgb]{0.753,0.753,0.753}}\textbf{0.814} & 0.844
    & 0.879 & {\cellcolor[rgb]{0.753,0.753,0.753}}\textbf{0.898}
    & {\cellcolor[rgb]{0.753,0.753,0.753}}\textbf{0.908} & 0.906
    & {\cellcolor[rgb]{0.753,0.753,0.753}}\textbf{0.909} \\
    \bottomrule
  \end{tabular}
\end{table}

\begin{figure}[tb]
  \centering
  \begin{subfigure}[b]{0.45\textwidth}
    \centering
    \includegraphics[width=\linewidth]{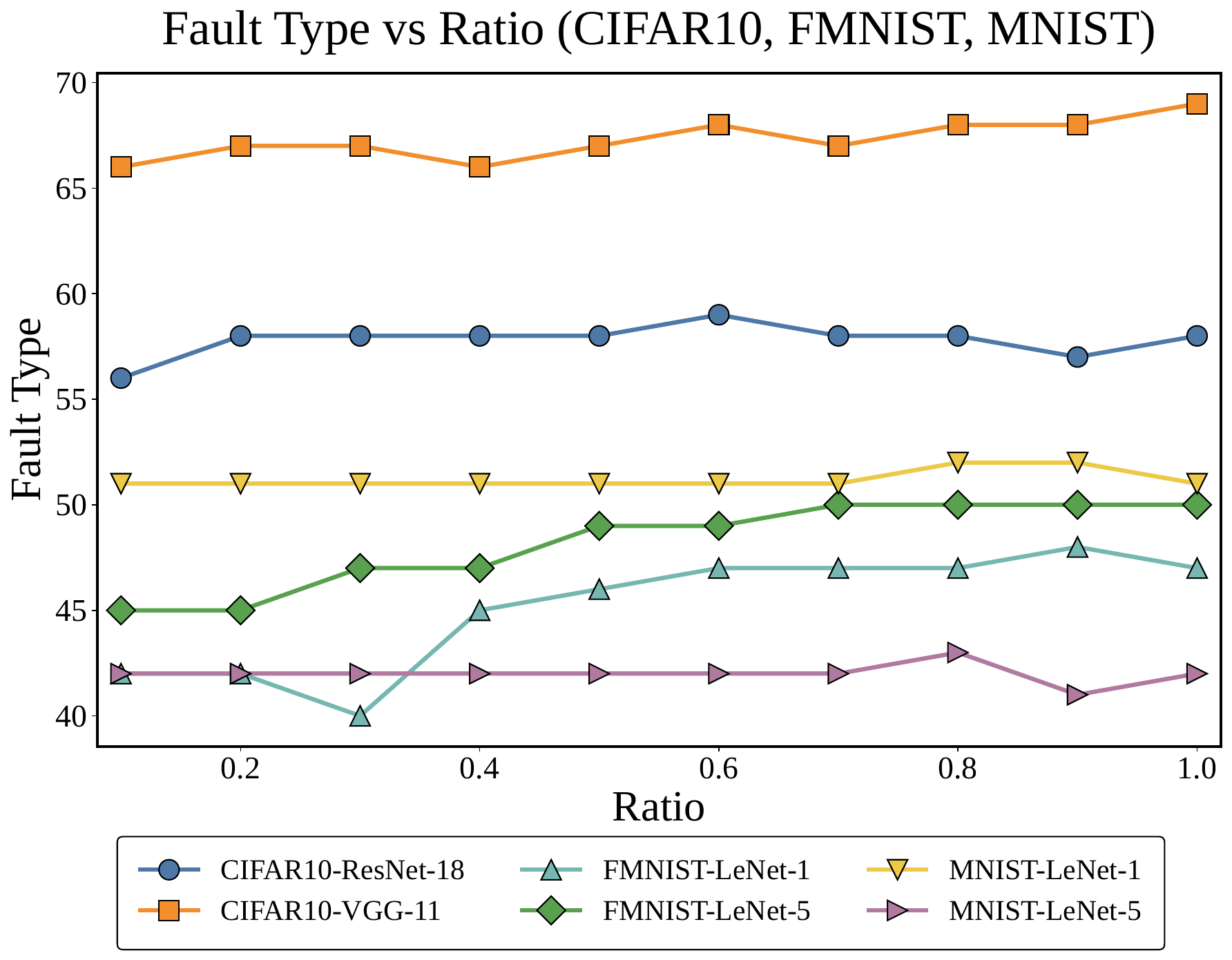}
  \end{subfigure} 
  \hspace{0.01\textwidth}
  \begin{subfigure}[b]{0.45\textwidth}
    \centering
    \includegraphics[width=\linewidth]{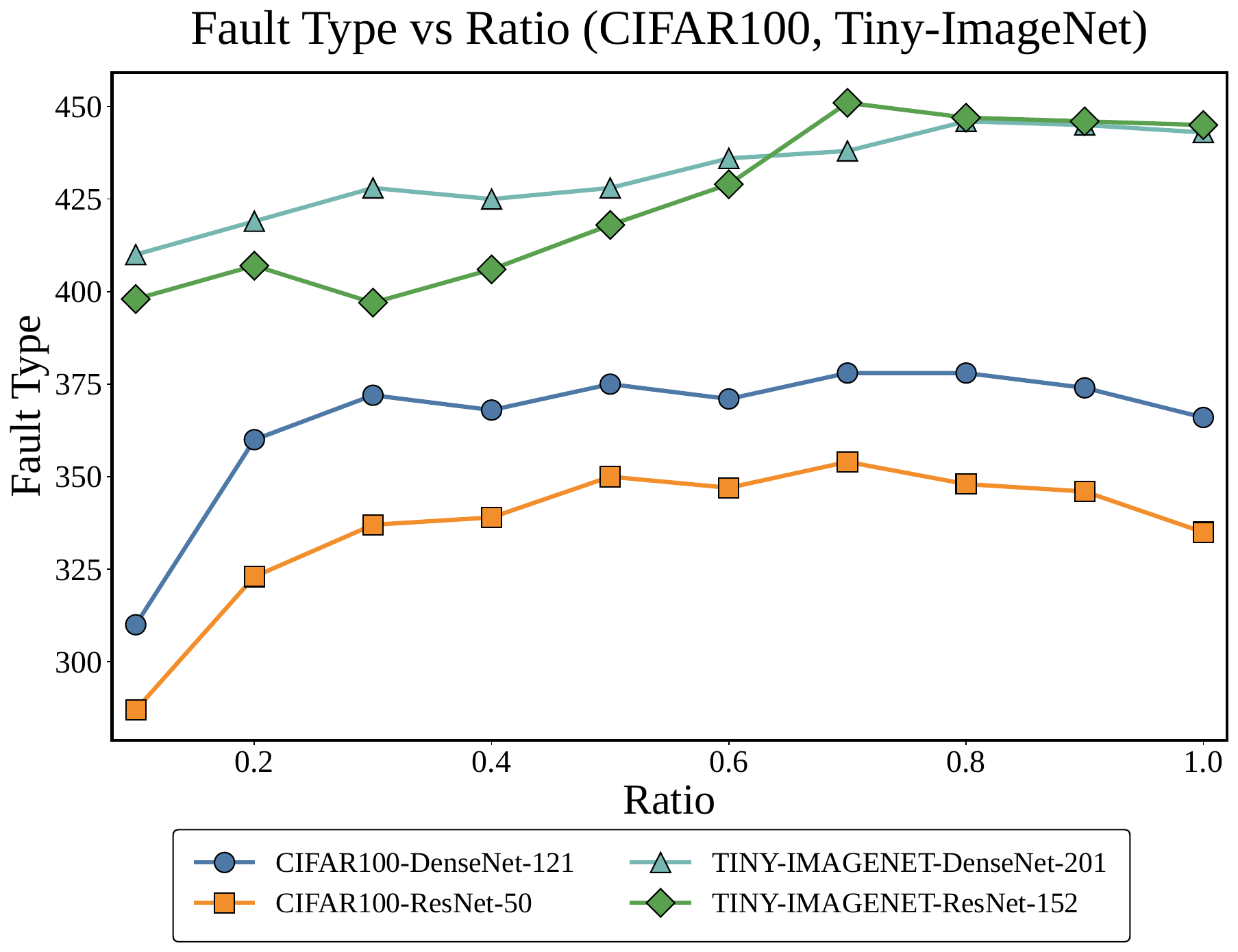}
  \end{subfigure} 
  \centering
  \begin{subfigure}[b]{0.45\textwidth}
    \centering
    \includegraphics[width=\linewidth]{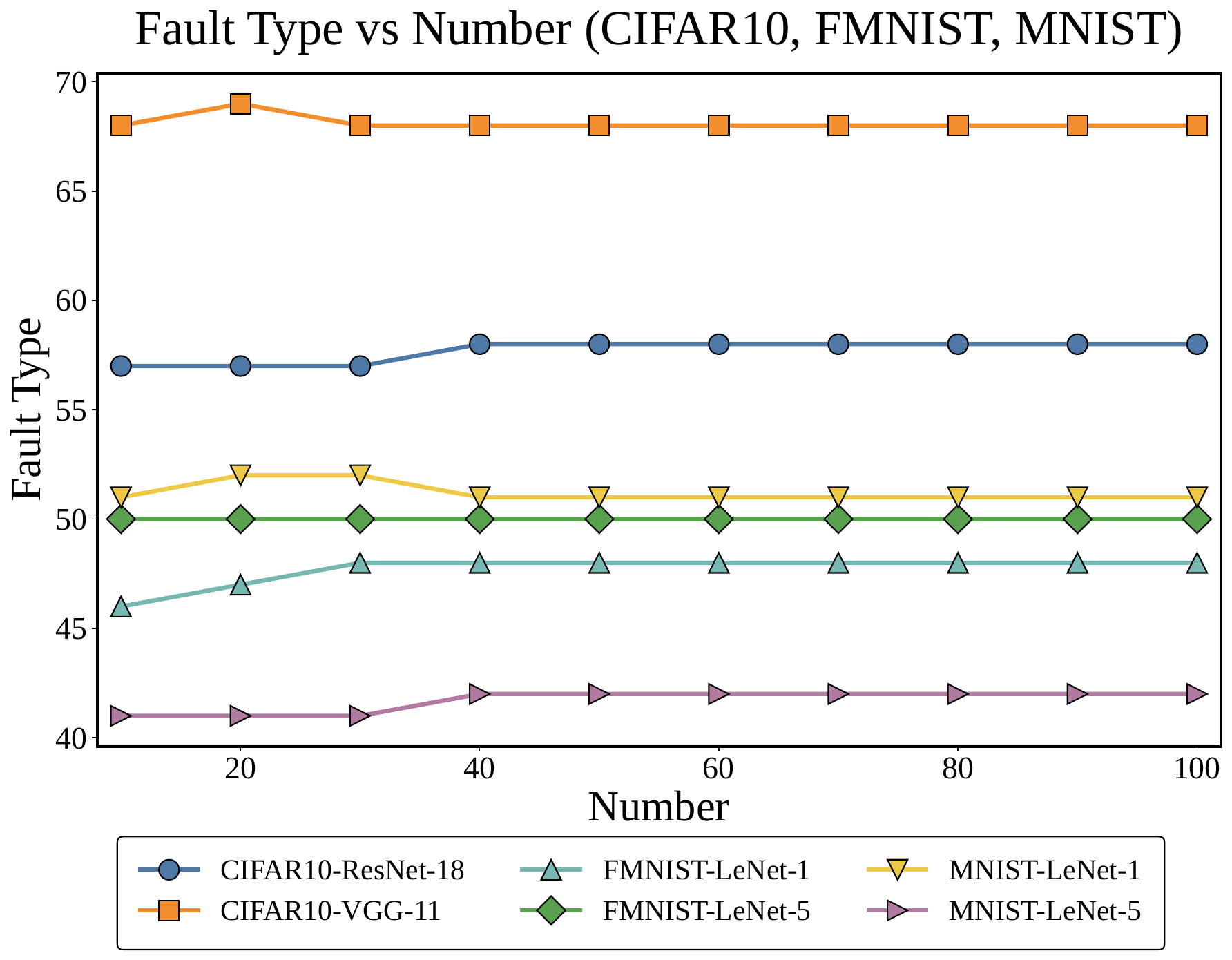}
  \end{subfigure} 
  \hspace{0.01\textwidth}
  \begin{subfigure}[b]{0.45\textwidth}
    \centering
    \includegraphics[width=\linewidth]{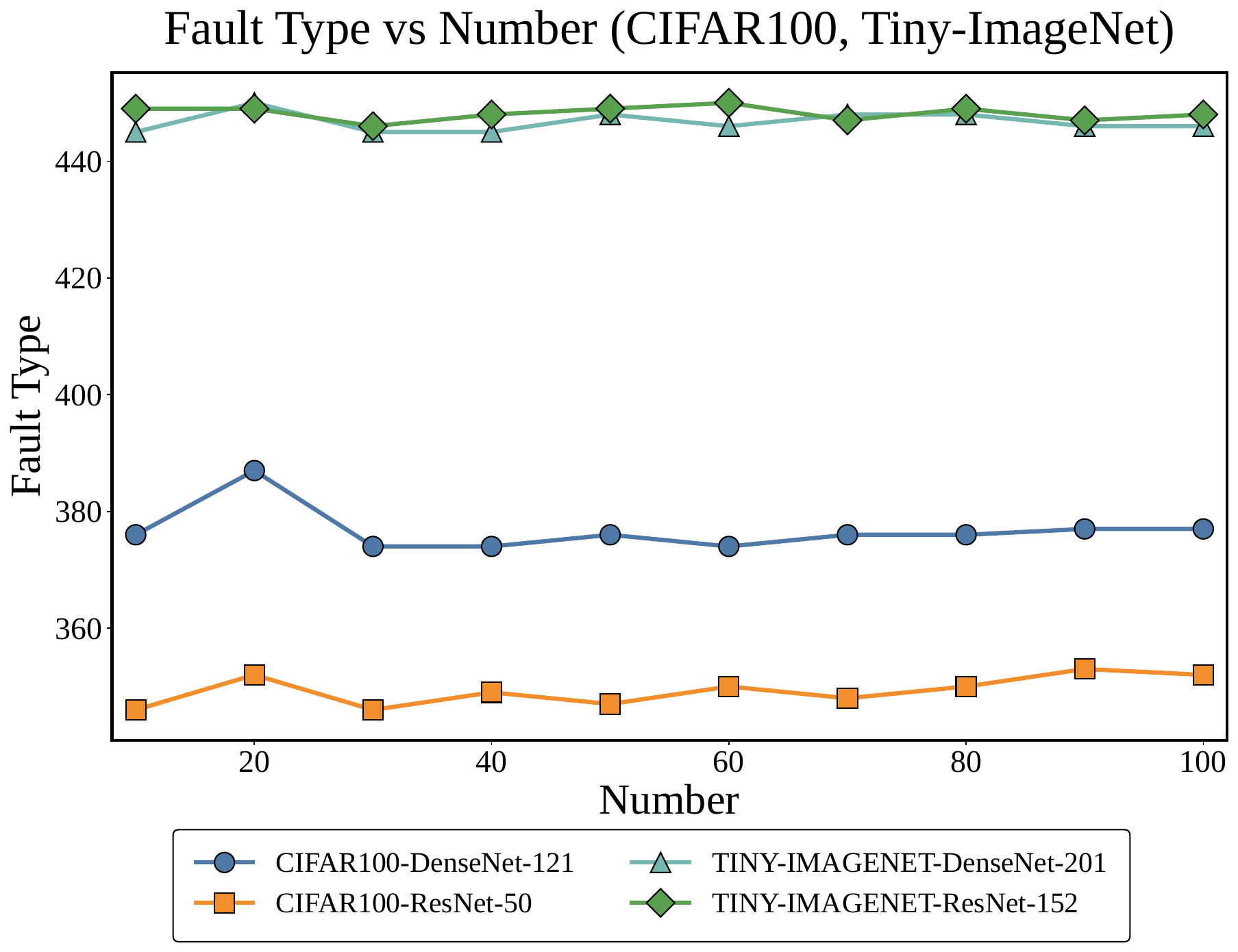}
  \end{subfigure} 
  \caption{The impact of the checkpoint selection ratio/number on fault diversity}
  \label{fig:The impact of the checkpoint selection ratio/number on fault diversity}
\end{figure}

The NC component can improve overall ranking quality in certain settings, most notably for transformer-based NLP models. On IMDB with Transformer, TVD+NC boosts RAUC-ALL from 0.715 to 0.851, suggesting that enforcing NC-guided structure provides a stronger global signal for uncertainty estimation in such architectures.

The full NCIP method consistently delivers the most robust performance across datasets and model families. By combining local boundary information with global geometric regularization, NCIP attains the best overall ranking quality; in particular, its average RAUC-ALL across all dataset-model pairs is the highest, reaching 0.909.

\begin{figure}[tb]
  \centering
  \includegraphics[width=0.90\linewidth]{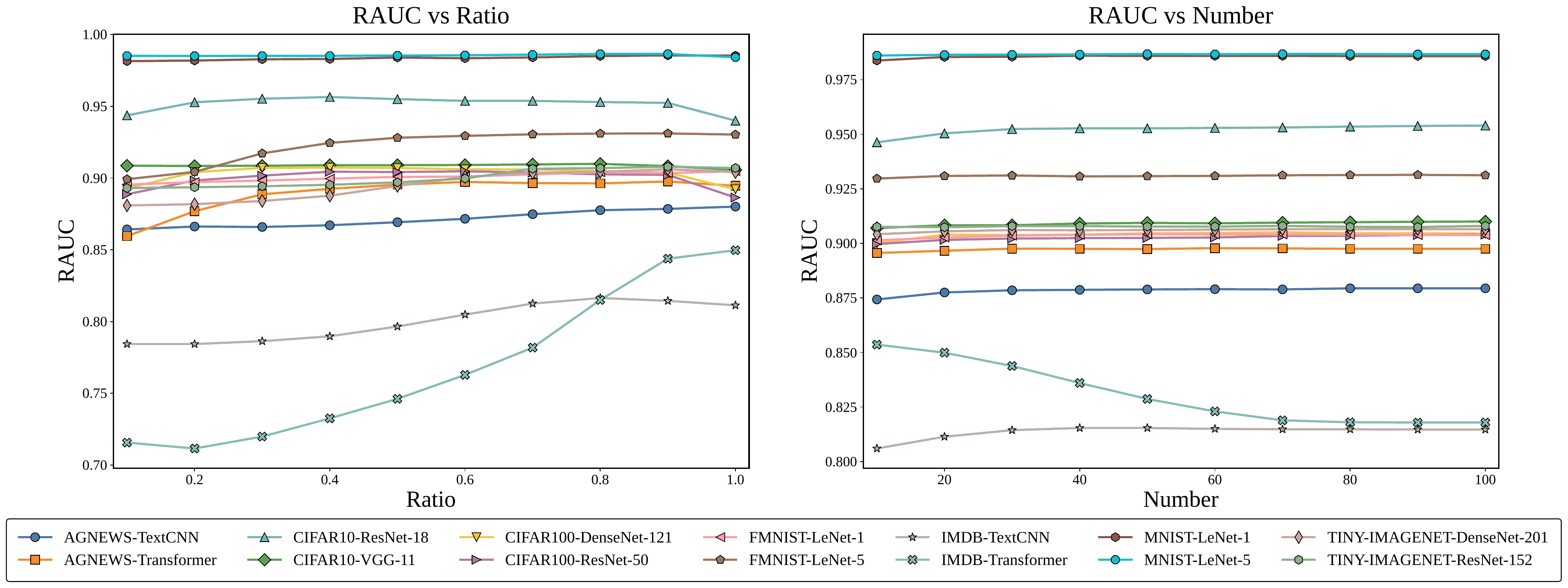}
  \caption{The impact of the checkpoint selection ratio/number on RAUC}
  \label{fig:The impact of the checkpoint selection ratio/number on RAUC}
\end{figure}

To examine hyperparameter sensitivity, we evaluate two factors: the checkpoint selection ratio $\alpha$ (candidate-pool ratio) and the checkpoint selection number $K$ (number of selected checkpoints). We report RAUC-ALL and fault diversity at a \SI{5}{\percent} budget, as shown in Fig.~\ref{fig:The impact of the checkpoint selection ratio/number on fault diversity} and Fig.~\ref{fig:The impact of the checkpoint selection ratio/number on RAUC}. $\alpha$ uses the last $\alpha$ fraction of checkpoints as candidates, and $K$ selects top-$K$ representatives from that pool.

For ratio sensitivity, increasing the candidate pool from $\alpha{=}0.1$ to about $0.7\!\sim\!0.9$ generally improves performance. On Tiny-ImageNet with DenseNet-201, diversity increases from 410 ($\alpha{=}0.1$) to 445 ($\alpha{=}0.9$). For number sensitivity, performance improves when $K$ is small and then plateaus; in practice, $K{\geq}30$ already gives near-saturated RAUC and diversity on most settings. Larger $K$ brings limited gains but increases multi-checkpoint inference cost.

Based on these results, we use $\alpha{=}0.9$ and $K{=}30$ as defaults, balancing effectiveness and efficiency.

\begin{figure}[tb]
  \centering
  \includegraphics[width=0.90\linewidth]{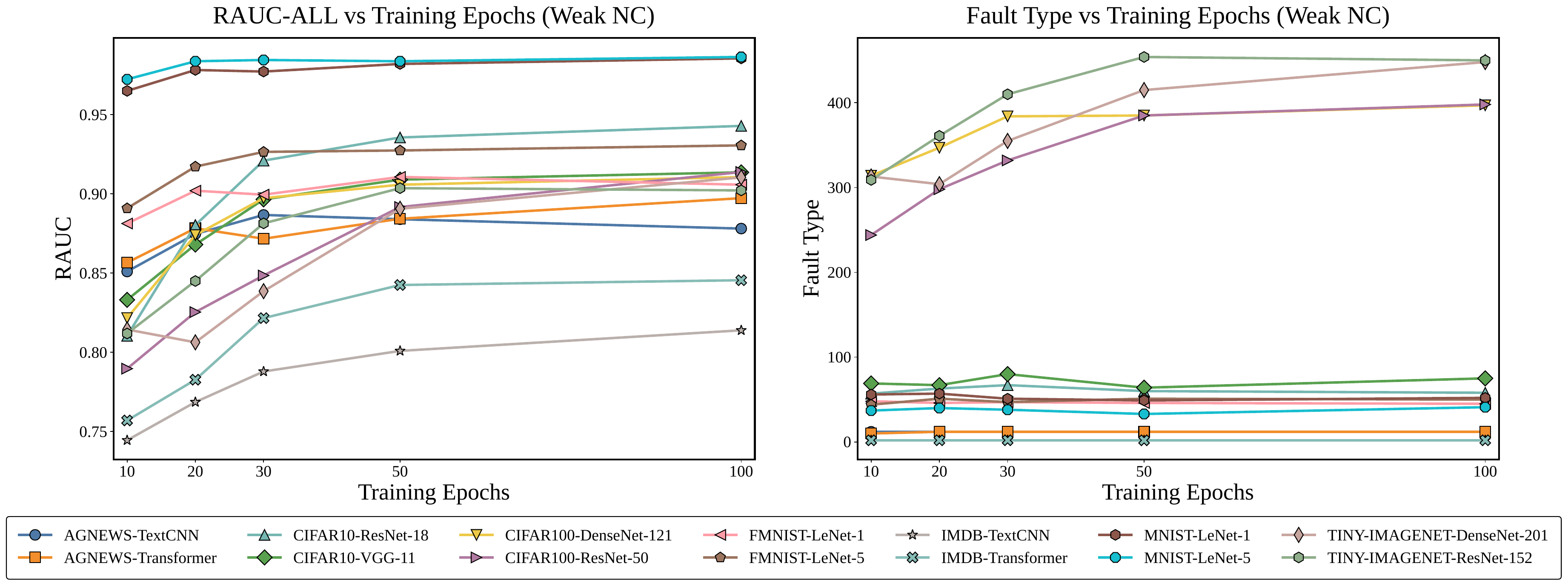}
  \caption{NCIP Performance under Weak Neural Collapse}
  \label{fig:ncip_performance_weak_nc}
\end{figure}

To assess performance under weak neural collapse, we report RAUC-ALL and fault type diversity across early training epochs in Fig.~\ref{fig:ncip_performance_weak_nc}. Since models from earlier training epochs exhibit weaker neural collapse, evaluating checkpoints from earlier training epochs approximates weaker NC conditions. RAUC-ALL generally increases as neural collapse becomes stronger. For example, CIFAR10-ResNet-18 improves from 0.8104 (10 checkpoints) to 0.9429 (100 checkpoints), and TinyImageNet-DenseNet-201 from 0.8144 to 0.9104. These results indicate that stronger neural collapse provides a more effective signal for test prioritization ranking.


\begin{tcolorbox}
    \noindent \textbf{Answer to RQ4: Ablation suggests that NC constraint and margin improve structure and boundary sensitivity, respectively. Hyperparameter analysis shows that increasing the checkpoint selection ratio improves RAUC-ALL and fault diversity up to about $\alpha=0.9$, while selection-number gains saturate around $K=30$.}
\end{tcolorbox}

\subsection{Efficiency (RQ5)}
\label{sec.Efficiency}
To evaluate computational efficiency, we report test case selection times in Table~\ref{tab:exec_time_methods}. For NCIP, the main overhead is in the selection stage, which includes multi-checkpoint loading and computing $\sigma_{\cos}$ for checkpoint filtering. This selection stage is a one-time cost per model. According to Table~\ref{tab:exec_time_methods}, NCIP-Select ranges from 1.96s to 306.87s across all settings, and remains much lower than training-heavy baselines such as TDPR-Train (up to 8003.94s) and EffiMAP-Train (up to 4371.56s). NCIP-Predict can still be higher on large models because inference uses multiple selected checkpoints (e.g., TinyImageNet), but the method keeps a practical effectiveness-runtime trade-off through controllable checkpoint budgeting.

\begin{table}
  \centering
  \caption{Execution time (in seconds) for each method. For clarity, we group LSA and DSA under SA, and aggregate uncertainty-based baselines (e.g., DeepGini, Entropy) under Uncert.}
  \label{tab:exec_time_methods}
  \scriptsize
  \setlength{\tabcolsep}{3.0pt}
  \renewcommand{\arraystretch}{1.2}
  \begin{tabular}{cc|ccccc|cc|cc|cc}
    \toprule
    \multirow{2}{*}{Dataset} & \multirow{2}{*}{Model}
    & \multirow{2}{*}{SA} 
    & \multirow{2}{*}{Uncert} 
    & \multirow{2}{*}{Dropout}
    & \multirow{2}{*}{NNS} 
    & \multirow{2}{*}{SETS}
    & \multicolumn{2}{c|}{EffiMAP}
    & \multicolumn{2}{c|}{TDPR}
    & \multicolumn{2}{c}{NCIP} \\
    & 
    &  &  &  &  &
    & Train & Predict
    & Train & Predict
    & Select & Predict \\
    \midrule

    \multirow{2}{*}{MNIST} 
      & LeNet-1
      & 78.15 & 5.46 & 6.38 & 8.92 & 67.19
      & 220.31 & 5.19
      & 70.70 & 4.58
      & 1.96 & 4.95 \\

      & LeNet-5
      & 84.13 & 5.13 & 8.07 & 7.86 & 65.46
      & 231.23 & 5.78
      & 1.88 & 9.43
      & 2.03 & 4.94 \\
    \midrule

    \multirow{2}{*}{Fashion-MNIST} 
      & LeNet-1
      & 75.34 & 5.57 & 7.55 & 7.70 & 69.23
      & 205.06 & 6.32
      & 185.16 & 4.88
      & 1.98 & 4.73 \\

      & LeNet-5
      & 72.93 & 5.70 & 7.20 & 8.82 & 67.58
      & 217.71 & 5.12
      & 156.13 & 4.96
      & 2.07 & 4.92 \\
    \midrule

    \multirow{2}{*}{CIFAR10} 
      & RN-18
      & 78.77 & 6.94 & 53.46 & 10.96 & 68.12
      & 654.02 & 6.99
      & 517.26 & 6.46
      & 19.91 & 33.57 \\

      & VGG-11
      & 82.84 & 6.37 & 12.84 & 11.06 & 74.35
      & 436.51 & 5.82
      & 484.17 & 5.92
      & 18.20 & 11.55 \\
    \midrule

    \multirow{2}{*}{CIFAR100} 
      & RN-50
      & 161.69 & 8.37 & 116.39 & 18.27 & 87.23
      & 912.30 & 8.40
      & 1881.14 & 7.46
      & 45.79 & 64.94 \\

      & DN-121
      & 140.41 & 7.82 & 90.60 & 12.56 & 78.10
      & 889.28 & 8.37
      & 1548.90 & 6.79
      & 27.79 & 57.62 \\
    \midrule

    \multirow{2}{*}{IMDB} 
      & CNN
      & 155.54 & 12.79 & 15.23 & 33.71 & 68.08
      & -- & --
      & 98.21 & 10.71
      & 9.56 & 13.95 \\

      & Trans.
      & 165.28 & 11.94 & 31.52 & 30.26 & 73.60
      & -- & --
      & 120.09 & 9.49
      & 11.00 & 24.94 \\
    \midrule

    \multirow{2}{*}{AGNEWS} 
      & CNN
      & 84.14 & 9.49 & 12.52 & 11.43 & 59.85
      & -- & --
      & 74.30 & 9.35
      & 30.65 & 10.91 \\

      & Trans.
      & 71.94 & 8.31 & 16.09 & 12.53 & 56.91
      & -- & --
      & 80.54 & 7.90
      & 9.19 & 12.01 \\
    \midrule

    \multirow{2}{*}{TinyImageNet} 
      & RN-152
      & 493.01 & 18.99 & 749.43 & 36.71 & 88.90
      & 4371.56 & 17.33
      & 8003.94 & 15.30
      & 148.31 & 374.21 \\

      & DN-201
      & 762.37 & 16.46 & 560.74 & 31.66 & 90.55
      & 3806.25 & 14.68
      & 5670.97 & 13.02
      & 306.87 & 311.10 \\
    \bottomrule
  \end{tabular}
\end{table}

\begin{tcolorbox}
  \noindent \textbf{Answer to RQ5: NCIP provides efficient prioritization with controllable overhead: the selection stage (multi-checkpoint loading and $\sigma_{\cos}$ computation) is required only once per model and is far cheaper than training-based baselines.}
\end{tcolorbox}

\section{Threats to Validity}
\textbf{Implementation and Evaluation Factors.} Our findings may be influenced by implementation and configuration factors. Any inaccuracies in checkpoint indexing, preprocessing consistency, or numerical stability in similarity/margin computations may bias the prioritization scores. In addition, training and evaluation stochasticity (e.g., random initialization, data shuffling, and GPU nondeterminism) can introduce variance. Moreover, adversarial robustness evaluation is sensitive to attack configurations; although we use widely adopted attacks (FGSM/PGD/CW/BIM) with standard library implementations and default settings, alternative hyperparameters may change the absolute RAUC values and, in some cases, relative rankings. To mitigate these threats, we enforce a unified training and evaluation protocol across all methods and use identical test sets and budgets. We also release our code and configuration to support reproducibility.

\textbf{Difference Between Model Disagreement-Based Methods.} A construct-validity threat is that different methods define model disagreement differently. Dropout and EffiMAP derive disagreement from stochastic perturbations around one checkpoint, whereas NCIP uses disagreement across NC-guided training checkpoints and combines it with final-margin information. Compared with disagreement-based methods that measure prediction variation through model perturbation, NCIP estimates prediction variability from naturally available training checkpoints. The two families rely on different sources of prediction instability, which may offer complementary perspectives in different deployment contexts.


\textbf{Practicality and Deployment Constraints.} NCIP incurs considerable runtime overhead due to loading checkpoints and performing multi-checkpoint inference to estimate prediction variability. This cost can hinder adoption in industrial pipelines where throughput and latency are critical. NCIP is mainly intended for model development and managed serving settings where training or fine-tuning checkpoints are retained, not for third-party deployed models without checkpoint access. Even with this overhead, NCIP is still useful in regression testing, where running every test in every cycle is often infeasible and reordering tests under a limited budget matters. This scope is also reflected in our LLM setting: we use only fine-tuning checkpoints (Fig.~\ref{fig:llm}) and NCIP remains competitive on WikiText (GPT-2 (0.587), OPT-125m (0.706)).


\section{Related Work}

\subsection{DNN Testing}
Testing DNN-based systems has been extensively studied~\cite{pei2017deepxplore, zheng2022neuronfair, harel2020neuron, guo2018dlfuzz, ma2018deepgauge, qiu2022detecting, you2023regression, hu2024test}. Since DNN development is largely data-driven~\cite{nielsen2015neural, alabdulmohsin2022revisiting}, a core challenge is to expose model failures effectively under constrained labeling, time, and computation budgets.

\subsection{Test Input Prioritization}
Test prioritization has become a particularly effective method \cite{al2022deepabstraction, wang2020dissector, hwang2023improved, li2023deeprank, huang2024neuron, li2025pricod, abbasishahkoo2025metasel, hu2025assessing}. This involves ranking test inputs based on their likelihood of failure, which allows for the early detection of critical errors within limited time and computational resources by focusing on high-risk samples. Existing DNN test-input prioritization techniques can be mainly divided into five categories: coverage-based methods, surprise adequacy-based methods, confidence-based methods, model-perturbation-based methods and learning-to-rank methods.

\textbf{Coverage-Based.} Coverage-based methods adapt traditional code-coverage ideas to DNNs by quantifying internal activation behaviors (e.g., neuron coverage) and prioritizing inputs that maximize such criteria~\cite{ma2018deepgauge, zhang2020neuron, guo2025white, yin2025lightweight}. Coverage-based methods focus on internal activation diversity, while NCIP focuses on prediction stability across the training trajectory.

\textbf{Surprise Adequacy-Based.} Surprise adequacy (SA) methods prioritize inputs by measuring how novel an input's activation trace is relative to the training distribution~\cite{kim2023evaluating, kim2020reducing, li2025improving}. Representative instantiations include KDE-based likelihood estimation and distance-based SA (DSA)~\cite{kim2019guiding}. SA has been used as a quantitative target to guide systematic boundary exploration~\cite{kang2024deceiving}. SA-based methods quantify input novelty relative to training distribution references. In contrast, NCIP operates on checkpoint-level prediction consistency without accessing training data at prioritization time.

\textbf{Confidence-Based.} These methods estimate error likelihood from output probabilities, e.g., DeepGini~\cite{feng2020deepgini}, entropy~\cite{byun2019input}, PCS~\cite{zhang2020towards}, and MSP~\cite{weiss2022simple}. Recent works improve uncertainty estimation by incorporating neighborhood uncertainty (NNS)~\cite{bao2023defense}, adjusting probability vectors via feature selection (FAST)~\cite{chen2024fast}, or jointly optimizing uncertainty and diversity (SETS)~\cite{wang2025sets}. Confidence-based methods derive ranking signals from per-sample output probabilities at a single training state. NCIP complements this perspective by measuring prediction variability across multiple training checkpoints, targeting instability that may not be captured by point-estimate confidence alone.

\textbf{Model-Perturbation-Based.} These methods prioritize inputs by measuring prediction changes under model-side perturbations. Dropout~\cite{hu2023aries} estimates uncertainty from repeated stochastic forward passes, while EffiMAP~\cite{wei2022predictive} perturbs both model and input and ranks samples by the resulting prediction variation. Model-perturbation methods estimate prediction variation through repeated perturbation at a fixed checkpoint. NCIP instead draws variation from naturally occurring training checkpoints, providing a different source of prediction instability.

\textbf{Learning-to-Rank.} Learning-to-rank methods learn a scoring function from labeled supervision so that bug-revealing inputs appear early under a budget. Examples include mutation-based labeling to improve prioritization (PRIMA)~\cite{wang2021prioritizing}, feature-based ranking for classical models (MLPrior)~\cite{dang2024test}, and approaches exploiting training dynamics to distinguish bug-revealing trajectories (TDPR)~\cite{shen2024prioritizing}. Learning-to-rank methods train a dedicated ranking model from labeled supervision, making effective use of failure information. NCIP is designed for settings where a trained model and its checkpoints are accessible but labeled test data is not, using cross-checkpoint variability as an unsupervised prioritization signal.

\section{Conclusion}
This paper presents \textbf{NCIP}, a label-free error prioritization approach for deep classifiers that leverages geometric stability during training. NCIP selects an NC-guided representative subset using classifier-weight equiangularity scores, and then prioritizes test inputs by combining (i) temporal prediction variability across these checkpoints with (ii) the final model prediction margin to capture decision boundary proximity. By tying misclassification risk to cross-checkpoint prediction variability, NCIP offers a principled alternative to confidence-only heuristics while requiring no training data during prioritization. Extensive experiments show that NCIP achieves competitive performance in early fault detection and delivers robust prioritization across datasets, architectures, and evaluation settings. Future work will explore extending NC-guided selection to broader model families and developing adaptive checkpoint selection strategies to further reduce computational overhead.

\begin{acks}
This work is supported by National Natural Science Foundation of China (Grant Nos. U25B2014, 62371069, 62372048, 62272056)
\end{acks}

\section*{Data Availability}
Our code is available at \url{https://github.com/lucky1207/NCIP}.

\section*{Declaration of Generative AI Use}
During the preparation of this work, the authors used GPT 5.2 in order to improve the readability and language of the manuscript.

\bibliographystyle{ACM-Reference-Format}
\bibliography{sample-base}



\end{document}